%% file: acl_latex.tex
\documentclass[11pt]{article}

\usepackage[final]{acl}

\usepackage{times}
\usepackage{latexsym}

\usepackage[T1]{fontenc}
\usepackage[utf8]{inputenc}
\usepackage{microtype}
\usepackage{inconsolata}
\usepackage{graphicx}

\usepackage{amsmath}
\usepackage{amssymb}
\usepackage{booktabs}
\usepackage{float}
\usepackage{listings}
\usepackage{multirow}
\usepackage{algpseudocode}
\usepackage{algorithm}
\usepackage{subcaption}

\title{Reward Design for Physical Reasoning in Vision-Language Models}

\author{Derek Lilienthal \\
  Pacific Northwest National Labratory \\\And
  Manisha Mukherjee \\
  Pacific Northwest National Labratory 
  \AND
  {\bf Sameera Horawalavithana} \\
  Pacific Northwest National Labratory \\}

\begin{document}
\maketitle

\begin{abstract}
\input{sections/0_abstract}
\end{abstract}

\section{Introduction}
\input{sections/1_Introduction}

\section{Related Work}
\input{sections/2_RelatedWork}

\section{Method}
\input{sections/3_Method}

\section{Experiments}
\input{sections/4_Experiments}

\section{Discussion}
\input{sections/5_Disucssion}

\section{Conclusion}
\input{sections/6_Conclusion.tex}

\bibliography{custom}

\appendix

\input{sections/7_Appendix}

\end{document}

%% file: sections/0_abstract.tex
Physical reasoning over images requires integrating perception, domain knowledge, and multi-step symbolic inference, yet Vision-Language Models (VLMs) fall far short of human performance on physics benchmarks. Reinforcement learning can improve reasoning via reward signals, but it remains unclear whether these signals must come from outside the model or can be derived from the model itself. We compare external rewards, which rely on an outside verifier (rule-based format and answer-correctness checks, or an LLM-judge rubric), with an internal attention-based reward computed from the model's own attention over image regions, requiring no labels or judge. 
We find that external rewards drive most of the accuracy gains, 
while preserving more coherent reasoning when rewarded for reasoning tokens. 
The internal attention-based reward, provides an annotation-free grounding signal that improves spatial-relation reasoning and is complementary to external rewards: adding it to format and accuracy rewards, at no additional annotation cost, improves average accuracy on Molmo~2 while generalizing to out-of-distribution tasks.
Our results suggest that internal attention rewards can supplement external supervision and lead to improved physical reasoning.

%% file: sections/1_Introduction.tex
Physical reasoning requires more than pattern recognition or factual recall. To solve a physics problem from an image, a model must decode implicit visual conditions, identify the governing physical laws, apply symbolic reasoning across multiple steps, and produce answers in correct units. This integration of perception and inference is fundamentally harder than mathematical reasoning or science knowledge retrieval \citep{phyx2025}. Current Vision Language Models (VLMs) reflect this difficulty. Even the strongest models achieve barely half the accuracy of human physics experts on rigorous multimodal benchmarks \citep{phyx2025}.
Supervised fine-tuning (SFT) has long been the standard post-training method for transforming base language models into assistants that produce helpful, structured responses. However, SFT maximizes the likelihood of a fixed reference output, which inherently rewards surface-level imitation rather than robust reasoning. Prior work confirms this training signal is effective for tasks dominated by memorization \citep{chu2025sft}, but it struggles to balance multiple objectives simultaneously and generalizes poorly to out-of-domain problems \citep{rlhf}.
Reinforcement learning has recently emerged as a powerful post-training paradigm for improving reasoning in language models. Group Relative Policy Optimization (GRPO) \citep{deepseekr1} has driven large gains by training models to maximize scalar reward signals without requiring a value network.

Nearly all of these rewards are \emph{external}: they depend on an outside verifier, such as ground-truth answers, rule-based checkers, or an LLM judge. 
Such supervision is costly to obtain in scientific domains, where labeled problems are scarce and judges are expensive and imperfect.
This is particularly useful for non- or weakly-verifiable tasks, which are difficult for humans to evaluate directly at scale.
For example, intuitive visual reasoning inherently risks subjective interpretation, and pure human expert annotation does not scale (expensive, slow, inter-rater variability on nuanced physical interpretations)~\citep{guo2026humanity}.
As these challenges motivate exploration of more general and scalable reward paradigms, \textit{Reinforcement Learning from Internal Feedback (RLIF)} was emerged as a paradigm in tuning models with internal feedback to improve performance without external rewards or supervision~\citep{zhao2026learning}. 

While there have been extensive GRPO‑based post‑training experiments on vision‑language models, they remain under‑explored for scientific models, particularly when it comes to exploring the reward‑design space. Prior work on process reward models and rubric-based feedback has explored structured external reward design for language tasks \citep{chen2026rm, gunjal2026rubrics, viswanathan2026checklists}, but these ideas have not been systematically studied in the multimodal physical reasoning setting, and none has been compared against rewards derived from the model itself.
Hence, in this study, we ask:
\textit{Can an internal reward derived from the model's own state improve physical reasoning in VLMs without labels, and how does it compare with, and complement, external rewards?}

We address this through a controlled comparison of external and internal rewards for GRPO-based VLM training on physical reasoning. We train two vision-language models of different scales, Gemma 4 E2B Instruction Tuned \citep{gemmateam2026gemma4} and Molmo2-O (7B) \citep{clark2026molmo2}, on the PhyX \citep{phyx2025} dataset, a 3,000-problem benchmark spanning six physics domains and six reasoning types with open-ended questions. The \textbf{external rewards} are \emph{verifiable} (output format and answer correctness against ground truth) and \emph{rubric-based} (LLM-judged physics principle identification, unit consistency, and reasoning quality). Our \textbf{internal reward}, the attention-based score (ASM), requires no annotation or judge, deriving its signal from the model's own attention weights over the input image. We evaluate each reward alone and in combination, against an SFT baseline, on PhyX and, zero-shot, on additional physics benchmarks.

Our main findings are that external rewards drive most of the accuracy gains, whereas ASM supplies a label-free grounding signal that improves spatial reasoning and is complementary to them: added on top of external rewards at no extra annotation cost, it improves accuracy on Molmo~2 while preserving gains on unseen benchmarks. Our main contributions are:
\vspace{-5pt}

\begin{itemize}
\item We introduce ASM, an internally-derived reward that leverages the model's own attention weights, providing a novel, label-free approach to reward design that improves the spatial reasoning of a vision language model.
\item A systematic comparison of external (verifiable and rubric-based) and internal (attention-based) rewards for VLM physical reasoning under GRPO, evaluated individually and in combination.
\item Empirical analysis across six physics domains and six reasoning types, revealing how each reward source shapes reasoning quality and structured output behavior.
\end{itemize}
\vspace{-10pt}

\begin{figure*}[!t]
    \centering
    \includegraphics[width=0.99\linewidth]{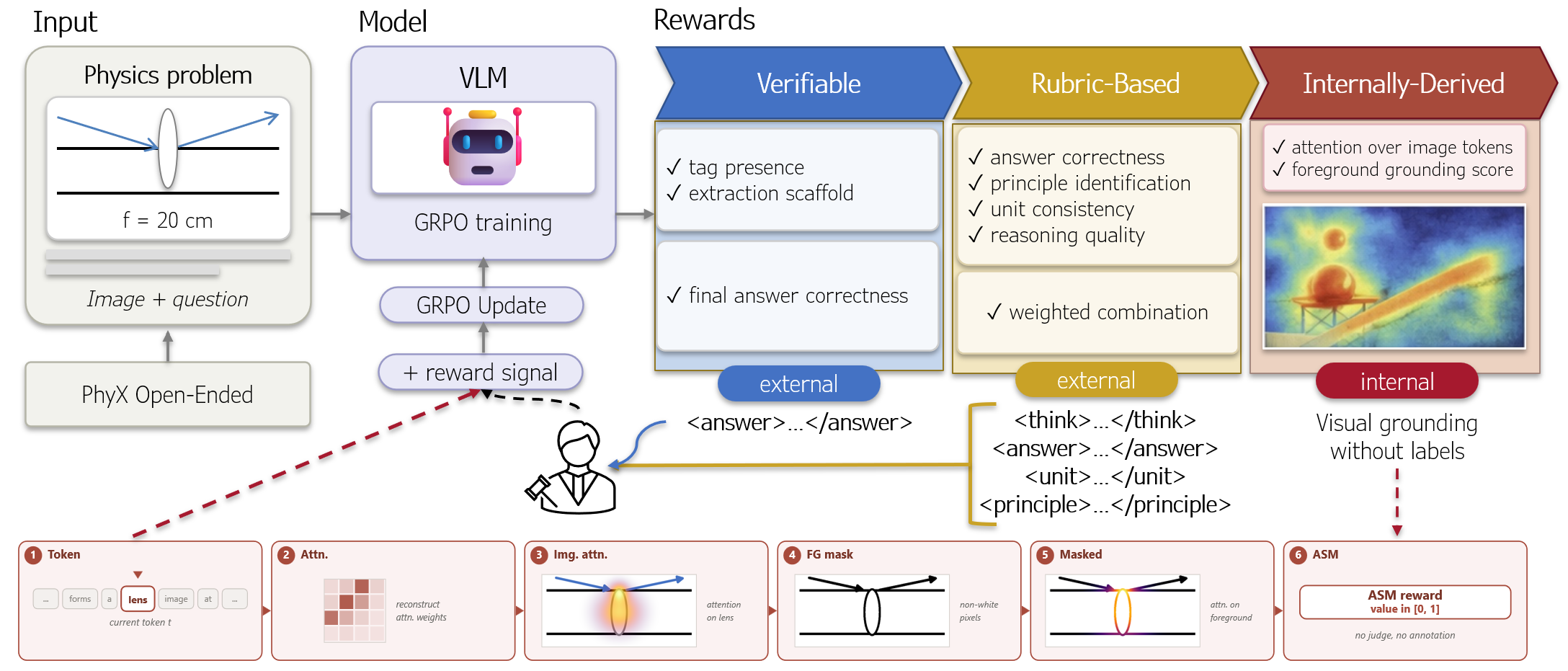}
   \caption{
\textbf{External and internal rewards for GRPO-based VLM training on physical reasoning.}
Given a PhyX image-question pair, the model is trained with GRPO using \textit{external} rewards (verifiable format and accuracy checks, and a rubric-based reward for correctness, principle, and units) or an \textit{internal} reward computed from its own attention over the image, which needs no labels or judge.
}
\vspace{-15pt}
    \label{fig:reward_design}
\end{figure*}

%% file: sections/2_RelatedWork.tex
\subsection{Vision-Language Models for Scientific Reasoning}
General-purpose VLMs such as LLaVA \citep{llava}, InternVL \citep{internvl}, and Qwen-VL \citep{qwenvl} have demonstrated strong performance on broad multimodal benchmarks but struggle on tasks requiring deep scientific understanding and reasoning \citep{horawalavithana2023scitune}. Science benchmarks such as ScienceQA \citep{scienceqa} and MMMU \citep{yue2023mmmu} largely test recall and shallow inference rather than multi-step physical reasoning grounded in realistic visual scenarios.
Existing physics benchmarks such as PHYBench \citep{phybench} and OlympiadBench \citep{olympiadbench} focus primarily on text-based or schematically simple problems, limiting assessment of multimodal reasoning. PhyX \citep{phyx2025} addresses these gaps with 3,000 visually-grounded problems spanning six physics domains and six reasoning types, in both multiple-choice and open-ended formats. 

\subsection{Reinforcement Learning for Multimodal Reasoning}
Reinforcement learning from human feedback (RLHF) \citep{rlhf} established reward-driven post-training as a powerful alignment tool. Subsequent work has adapted it specifically to improve reasoning. DeepSeek-R1 \citep{deepseekr1} introduced GRPO, which replaces the value network with group-relative advantage estimation. This yields strong gains on mathematical reasoning without the instability of PPO. Extensions to VLMs are emerging \citep{shen2025vlm,wang2025vl}, but systematic study of how reward design affects VLM reasoning in scientific domains remains limited.

Reward designs for reasoning broadly fall into three paradigms. \textit{Verifiable rewards}, exemplified by DeepSeek-R1-style exact-match or rule-checked correctness, supervise a single ground-truth-checkable signal and require no learned judge. \textit{Rubric-based rewards} score multiple structured dimensions of an output simultaneously, extending the denser, step-level supervision of process reward models (PRMs) \citep{prm} and rubric-based feedback \citep{rubric} to richer, partly LLM-judged criteria beyond a single correctness signal. \textit{Internally-derived rewards} instead compute the training signal from the model's own internal state rather than its output, requiring no external annotation or judge. Recent work encourages VLMs to reason more with images using Intersection-over-Union (IoU) rewards \citep{liu2025visual}, where the VLM predicts object locations in text that are compared against ground-truth pixel labels; this is a verifiable reward over a spatial output and requires bounding-box annotations.
In our work, we directly leverage the VLM's internal attention weights during token generation as an internally-derived reward, with no spatial annotations required.
Rather than supervising what the model says about image locations, we supervise where the model attends while generating its response.

%% file: sections/3_Method.tex
 
\subsection{Problem Formulation}
We formulate physical reasoning as a conditional generation task. Given an image $\mathbf{v}$ and a
question $q$, the model generates a structured response $y = (c, a, u, p)$ comprising a
chain-of-thought $c$, a final answer $a$, a physical unit $u$, and an identified physical principle
$p$. We train using GRPO~\citep{deepseekr1}, which optimizes the policy $\pi_\theta$ by sampling a
group of $G$ completions per prompt and computing advantages relative to the group mean reward,
without requiring a separate value network. The training objective is:
{\setlength{\abovedisplayskip}{3pt}\setlength{\belowdisplayskip}{3pt}\setlength{\abovedisplayshortskip}{0pt}\setlength{\belowdisplayshortskip}{3pt}%
\begin{multline}
    \mathcal{L}_{\text{GRPO}}(\theta) = -\mathbb{E}\left[\sum_{i=1}^{G} \hat{A}_i \log \pi_\theta(y_i \mid \mathbf{v}, q)\right] \\ + \beta \, D_{\text{KL}}(\pi_\theta \| \pi_{\text{ref}})
\end{multline}}
where $\hat{A}_i = (r_i - \bar{r}) / \sigma_r$ is the normalized group advantage and $\beta$
controls the KL penalty against the reference policy $\pi_{\text{ref}}$.
 
\subsection{Reward Signal Designs}
\label{sec:rewards}
All reward functions share a common structure: per-sample scalar rewards are computed over a group
of $G = 8$ completions per prompt, then normalized by subtracting the group mean to form
advantages. 
The model is prompted to produce outputs structured with four tags: \texttt{<think>}, \texttt{<answer>}, \texttt{<unit>}, and \texttt{<principle>}. We compare two sources of reward. \textbf{External} rewards are specified by an outside verifier: \emph{verifiable} rewards (rule-based format and judge-checked answer accuracy) and a \emph{rubric-based} reward scoring multiple judge-graded dimensions. The \textbf{internal} reward is computed entirely from the model's own attention, requiring no labels or judge.
We denote rewards by their terms: rule-based format ($r_f$), LLM-judge (jury) answer accuracy ($r_a$), LLM-judge (rubric) answer and physics reasoning accuracy ($r_{\text{rub}}$), and the attention-mask score (ASM). Configurations are sums of these terms, and the result tables use the same notation; in the text we abbreviate $r_f$, $r_f + r_a$, and $r_{\text{rub}}$ as Fmt, Fmt+Acc, and Rubric.

\subsubsection{External Reward: Verifiable}

\paragraph{Format ($r_f$).}
The format reward $r_f$ supervises output structure only, independent of content correctness: each of the four required tag pairs (\texttt{think}, \texttt{answer}, \texttt{unit}, \texttt{principle}) that is present contributes $1/4$, so $r_f \in [0,1]$.
 
\paragraph{Accuracy ($r_a$).}
The accuracy reward $r_a$ supervises answer correctness only. Answer correctness is assessed by an LLM judge, GPT-oss 120B~\citep{gptoss}, queried as a 3-judge jury where each judge scores correctness on a 0/1/2 scale (incorrect, partially correct, fully correct), and scores are averaged across judges, allowing partial credit for approximately correct answers.
 
\subsubsection{External Reward: Rubric-Based}
The rubric-based reward $r_{\text{rub}}$ extends verifiable accuracy supervision with three further judge-graded, physics-grounded components. For open-ended questions, rule-based matching is brittle to scientific notation, unit prefixes, numerical precision, and free-text explanations, so we combine the format reward $r_f$ with a structured LLM judge call that jointly scores all rubric dimensions. We query the same judge (GPT-oss 120B) $K = 3$ times per completion, averaging the continuous dimensions (correctness, reasoning) and taking a majority vote for the binary ones (unit, principle). The four dimensions are \textit{correctness} $\in \{0,1,2\}$ (incorrect, partially correct, fully correct; $r_a = \text{correctness}/2$), \textit{principle} $\in \{0,1\}$ (the correct physics law is applied), \textit{unit} $\in \{0,1\}$ (the stated unit is correct), and \textit{reasoning} $\in \{0,1,2\}$ (no, partial, or valid step-by-step derivation; $r_{\text{reason}} = \text{reasoning}/2$). The judge may not award full correctness credit if the reasoning is invalid, so correct answers reached by spurious means are not rewarded.

\textit{Combined rubric reward.}
The rubric combines its components with a weighted sum:
{\setlength{\abovedisplayskip}{3pt}\setlength{\belowdisplayskip}{3pt}\setlength{\abovedisplayshortskip}{0pt}\setlength{\belowdisplayshortskip}{3pt}%
\begin{multline}
    r_{\text{rub}} = 0.50 \cdot r_a + 0.15 \cdot r_p + 0.10 \cdot r_u \\ + 0.15 \cdot r_{\text{reason}} + 0.08 \cdot r_f
\end{multline}}
A soft penalty multiplies the total by $0.6$ when reasoning is absent ($r_{\text{reason}} = 0$), a small length penalty $\min(\lvert y \rvert / 4000,\, 0.05)$ discourages verbose outputs, and all rewards are clipped to $[0, 1]$.

\subsubsection{Internal Reward: Attention Score Mask (ASM)}
The attention-weight reward $r_{\text{attn}}$ is our internal reward: it is computed from the model's own attention distribution over image tokens, requiring no annotation, external model, or judge, and encourages the model to visually ground its reasoning in the foreground content of the input image.

\textit{Attention extraction.}
We register forward hooks on the query and key projection layers ($\mathbf{W}_Q$, $\mathbf{W}_K$) of the final transformer layer and the rotary embedding module. We then manually reconstruct the full attention matrix from captured projections:
{\setlength{\abovedisplayskip}{3pt}\setlength{\belowdisplayskip}{3pt}\setlength{\abovedisplayshortskip}{0pt}\setlength{\belowdisplayshortskip}{3pt}%
\begin{equation}
    \mathbf{A} = \text{softmax}\!\left(\mathbf{Q}_{\text{rope}} \mathbf{K}_{\text{rope}}^\top \cdot \alpha \cdot \mathbf{M}_{\text{causal}}\right)
\end{equation}}
where $\alpha$ is the model-specific attention scaling factor and $\mathbf{M}_{\text{causal}}$ is the causal mask.
Grouped-query attention heads are expanded via repeat-interleave before the dot product.
The resulting attention tensor is averaged across all heads, and the final token's attention row is extracted, yielding a per-token attention vector over the full input sequence for each rollout.
This reconstruction approach avoids quadratic memory growth from storing raw attention weights during the forward pass.

\textit{Image attention map.} Attention over image tokens is reshaped into a 2D grid ($\frac{H_{\text{img}}}{p} \times \frac{W_{\text{img}}}{p}$), then normalized and resized to the original image resolution.

\textit{Foreground grounding score.} The reward is the fraction of attention mass on non-white foreground pixels. A binary mask $\mathbf{F}$ is constructed by thresholding pixels (RGB channels $\geq 230$ = background):
{\setlength{\abovedisplayskip}{3pt}\setlength{\belowdisplayskip}{3pt}\setlength{\abovedisplayshortskip}{0pt}\setlength{\belowdisplayshortskip}{3pt}%
\begin{equation}
\label{eq:r_attn}
    r_{\text{attn}} = \frac{\sum_{i,j} \hat{\mathbf{A}}_{ij} \cdot \mathbf{F}_{ij}}{\sum_{i,j} \mathbf{F}_{ij}}
\end{equation}}
where $\hat{\mathbf{A}}$ is the resized, normalized attention map. This yields a reward in $[0, 1]$ that is high when the model attends to meaningful image content and low when attention is diffuse or on whitespace. To obtain a single scalar over the full generated sequence, we average Equation~\ref{eq:r_attn} across all $T$ generated tokens and define the result as the \textbf{Attention Score Mask (ASM)}. This averaged score serves as the reward signal during GRPO training. See Algorithm~\ref{alg:attention-mask-reward} in the Appendix for implementation details.
{\setlength{\abovedisplayskip}{3pt}\setlength{\belowdisplayskip}{3pt}\setlength{\abovedisplayshortskip}{0pt}\setlength{\belowdisplayshortskip}{3pt}%
\begin{equation}
\label{eq:asm}
    \text{ASM} = \frac{1}{T} \sum_{t=1}^{T} r_{\text{attn}}^{(t)}
\end{equation}}

Because ASM depends only on the model and the input image, it composes with the external rewards by simple addition (e.g., $r_f + r_a + \text{ASM}$, denoted Fmt+Acc+ASM), contributing a grounding signal at no additional annotation cost.

%% file: sections/4_Experiments.tex
\begin{table*}[ht!]
\centering
\small
\resizebox{\textwidth}{!}{%
\begin{tabular}{lccccccccccccccc}
\toprule
 &  & \multicolumn{7}{c}{PhyX} & \multicolumn{6}{c}{MMMU} & \multicolumn{1}{c}{SciBench} \\
\cmidrule(lr){3-9}\cmidrule(lr){10-15}\cmidrule(lr){16-16}
Method & Avg & P-Ovr & P-EM & P-Mech & P-Mod & P-Opt & P-Th & P-W/A & M-Ovr & M-CM & M-EM & M-NP & M-Opt & M-TSM & S-Fund \\
\midrule
\multicolumn{1}{c}{\textbf{Gemma-4-E2B-it}} \\
Baseline & 0.066 & 0.047 & 0.077 & 0.073 & 0.049 & 0.023 & 0.023 & 0.036 & 0.110 & 0.104 & 0.096 & 0.167 & 0.187 & 0.057 & 0.060 \\
SFT & 0.057 & 0.049 & 0.069 & 0.062 & 0.045 & 0.036 & 0.041 & 0.040 & 0.072 & 0.069 & 0.073 & 0.130 & 0.053 & 0.076 & 0.044 \\
GRPO ($r_f$) & 0.101 & 0.071 & 0.067 & 0.103 & 0.065 & 0.046 & 0.060 & 0.085 & 0.152 & 0.143 & 0.136 & 0.167 & 0.200 & 0.181 & 0.082 \\
GRPO ($r_f$ + ASM) & 0.080 & 0.053 & 0.048 & 0.079 & 0.049 & 0.038 & 0.037 & 0.067 & 0.127 & 0.119 & 0.108 & 0.185 & 0.167 & 0.162 & 0.060 \\
GRPO ($r_f + r_a$) & \textbf{0.209} & \textbf{0.183} & \textbf{0.131} & \textbf{0.284} & \textbf{0.203} & \textbf{0.137} & \underline{0.133} & \underline{0.204} & \textbf{0.255} & \textbf{0.268} & 0.202 & \textbf{0.444} & \textbf{0.287} & 0.257 & \underline{0.213} \\
GRPO ($r_f + r_a$ + ASM) & \underline{0.201} & 0.171 & \underline{0.127} & 0.268 & 0.169 & 0.124 & \textbf{0.143} & 0.192 & \underline{0.251} & \underline{0.259} & \textbf{0.221} & 0.352 & 0.240 & \underline{0.295} & 0.191 \\
GRPO ($r_{\text{rub}}$) & 0.200 & \underline{0.178} & 0.125 & \underline{0.278} & \underline{0.193} & \underline{0.133} & 0.112 & \textbf{0.220} & 0.236 & 0.223 & \underline{0.204} & \underline{0.407} & \underline{0.247} & \textbf{0.324} & \textbf{0.219} \\
\midrule
\multicolumn{1}{c}{\textbf{Molmo2-O-7B}} \\
Baseline & 0.082 & 0.057 & 0.062 & 0.077 & 0.051 & 0.038 & 0.046 & 0.065 & 0.144 & \textbf{0.160} & 0.122 & \textbf{0.352} & 0.100 & 0.105 & 0.022 \\
SFT & 0.077 & 0.056 & \underline{0.071} & 0.063 & 0.063 & 0.034 & 0.035 & 0.065 & 0.125 & 0.134 & 0.113 & \underline{0.278} & 0.120 & 0.057 & 0.016 \\
GRPO ($r_f$) & 0.061 & 0.045 & 0.058 & 0.054 & 0.030 & 0.030 & 0.031 & 0.067 & 0.087 & 0.089 & 0.073 & 0.222 & 0.080 & 0.076 & 0.093 \\
GRPO ($r_f$ + ASM) & 0.062 & 0.045 & 0.050 & 0.052 & 0.049 & 0.023 & 0.052 & 0.046 & 0.098 & 0.101 & 0.077 & 0.204 & 0.093 & 0.124 & 0.060 \\
GRPO ($r_f + r_a$) & \underline{0.107} & \underline{0.085} & 0.058 & \underline{0.131} & \underline{0.065} & \underline{0.061} & \underline{0.066} & \underline{0.129} & \underline{0.147} & 0.145 & \underline{0.138} & 0.222 & \textbf{0.147} & \underline{0.152} & \textbf{0.148} \\
GRPO ($r_f + r_a$ + ASM) & \textbf{0.115} & \textbf{0.097} & \textbf{0.075} & \textbf{0.139} & \textbf{0.073} & \textbf{0.084} & \textbf{0.077} & \textbf{0.133} & \textbf{0.155} & \underline{0.151} & \textbf{0.143} & 0.185 & \underline{0.133} & \textbf{0.238} & \underline{0.142} \\
GRPO ($r_{\text{rub}}$) & 0.058 & 0.049 & 0.014 & 0.091 & 0.037 & 0.038 & 0.041 & 0.073 & 0.073 & 0.067 & 0.070 & 0.019 & 0.113 & 0.086 & 0.060 \\
\bottomrule
\end{tabular}%
}
\caption{Mean judge score on PhyX (P), MMMU (M) and SciBench (S) for baseline and SFT models; scores averaged over 3 runs of each config. Within each model, the best score per column is in bold and the second best underlined. GRPO rows are labeled by their reward terms: $r_f$ = rule-based format reward, $r_a$ = LLM-judge answer accuracy, $r_{\text{rub}}$ = LLM-judge (rubric) answer and physics reasoning accuracy, ASM = attention-mask score. \textbf{P-} = PhyX: Overall (Ovr), Electromagnetism (EM), Mechanics (Mech), Modern Physics (Mod), Optics (Opt), Thermodynamics (Th), Waves and Acoustics (W/A). \textbf{M-} = MMMU: Overall (Ovr), Classical Mechanics (CM), Electromagnetism (EM), Nuclear Physics (NP), Optics (Opt), Thermodynamics and Statistical Mechanics (TSM). \textbf{S-} = SciBench: Fundamentals of Physics (Fund). Avg = weighted average of the datasets' overall scores.}
\label{tab:oe-sft-avg}
\end{table*}

\subsection{Experimental Setup}

\paragraph{Models.} We use Gemma 4 E2B-it and Molmo2-O (7B), two instruction-tuned VLMs that differ in scale (5B--7B effective parameters) and architecture, each trained and evaluated independently under the same reward configurations (hyperparameters and hardware in Appendix~\ref{app:training_setup}).

\paragraph{Data.} We train on 2,000 open-ended PhyX problems and evaluate on its disjoint testmini subset (1,000 problems). Each example provides the question, image, ground-truth answer, domain, and physics subfield (a coarse principle hint for the rubric judge); we add ground-truth unit labels (Appendix~\ref{appendix:dataset_statistics}) and show the prompts in Appendix~\ref{app:examples}. We also evaluate zero-shot on an open-ended version of the MMMU physics subset~\citep{yue2023mmmu} and the SciBench fundamental physics subset~\citep{wang2024scibench}.

\paragraph{Evaluation.} Each configuration, including the SFT baseline, is trained 3 times; we sample 5 responses per question from each trained model and report avg@15. An LLM judge scores a response as correct if its \texttt{<answer>} is semantically equivalent to the ground truth. Since GPT-oss 120B computes the accuracy and rubric rewards during training, scoring with it could favor the optimized checkpoints, so we re-score every response with Claude Haiku~5.5~\citep{anthropic2026claudehaiku55}, a judge from a different model family that was never used as a reward, and report only these scores (details and agreement with GPT-oss in Appendix~\ref{app:final_judge}).

\subsection{Results}

\paragraph{External vs.\ Internal Rewards.} 

Table~\ref{tab:oe-sft-avg} reports accuracy for the untrained baseline, SFT, and the GRPO reward configurations (standard deviations over runs are in Table~\ref{tab:oe-sft-avg-std} in the Appendix). The answer-accuracy reward gives the clearest PhyX gains: $r_f + r_a$ raises accuracy from 0.047 to 0.183 on Gemma~4 and from 0.057 to 0.085 on Molmo~2, whereas SFT does not help (0.049 and 0.056) and format-only rewards help little ($r_f$: 0.071 and 0.045). The internal ASM reward is most useful alongside an accuracy reward: adding it to $r_f + r_a$ requires no extra labels or judge, gives the best PhyX accuracy on Molmo~2 (0.097 vs.\ 0.085), and stays close on Gemma~4 (0.171 vs.\ 0.183), both within one standard deviation across runs, whereas $r_f$ + ASM is no better than $r_f$ alone (0.053 and 0.045). The rubric reward $r_{\text{rub}}$ is close to $r_f + r_a$ on Gemma~4 (0.178 vs.\ 0.183) but well below it on Molmo~2 (0.049 vs.\ 0.085, with large run-to-run variability, $\pm$0.043).

\paragraph{Out-of-distribution Generalization: MMMU and SciBench}
\label{sec:molmo2_generalization}

We also evaluate the PhyX-trained checkpoints zero-shot on the physics subset of MMMU and the Fundamentals of Physics subset of SciBench (M-* and S-Fund in Table~\ref{tab:oe-sft-avg}), using the same prompts, judge, and avg@15 protocol; benchmark details are in Appendix~\ref{app:ood_benchmarks}.

\paragraph{Findings.} \emph{The external accuracy reward transfers, the internal attention reward retains those gains at no extra annotation cost, and the rubric reward $r_{\text{rub}}$ does not add to them.} GRPO with $r_f + r_a$ transfers: it improves SciBench for both models (from 0.060 to 0.213 on Gemma~4 and from 0.022 to 0.148 on Molmo~2) and MMMU for Gemma~4 (from 0.110 to 0.255), but not MMMU for Molmo~2 (0.144 to 0.147), where no configuration clearly exceeds the baseline. Adding ASM preserves this transfer: MMMU 0.251 vs.\ 0.255 on Gemma~4 and 0.155 vs.\ 0.147 on Molmo~2, and SciBench 0.191 vs.\ 0.213 and 0.142 vs.\ 0.148, all within one standard deviation, so the in-domain gain of ASM on Molmo~2 does not come at the expense of out-of-distribution accuracy. The rubric reward $r_{\text{rub}}$ is below $r_f + r_a$ on MMMU for both models (0.236 vs.\ 0.255 on Gemma~4, within the run-to-run variability, and 0.073 vs.\ 0.147 on Molmo~2) and on par with it on SciBench for Gemma~4 (0.219 vs.\ 0.213).

%% file: sections/5_Disucssion.tex
In this section, we discuss the performance of the GRPO-trained models under different reward configurations. 

\paragraph{Does the attention reward contribute to visual reasoning tasks?}

We examine open-ended results from two angles: PhyX accuracy by reasoning type (Table~\ref{tab:reasoning-types}), and the concentration of attention over the image, measured by the normalized entropy of the cumulative attention map (Ent.; Appendix~\ref{appendix:attention_entropy_score}) on the PhyX test, MMMU, and SciBench questions for the best checkpoint of each configuration. GRPO with $r_f + r_a$ improves most reasoning types (all six on Gemma~4, five of six on Molmo~2), whereas SFT gives no consistent gain. The contribution of ASM on top of $r_f + r_a$ is model-dependent. On Molmo~2 it improves five of six types, most clearly Spatial Relation (0.108 to 0.135) and Implicit Condition (0.069 to 0.109), and gives the highest overall accuracy of any Molmo~2 configuration (0.097) without extra annotations or judge calls. It also lowers attention entropy (0.773 to 0.683, lower on 71\% of questions and on every benchmark): attention is concentrated on fewer image patches, so the non-masked (foreground) content receives more of it. On Gemma~4, ASM lowers five of six types (improving only Implicit Condition, 0.132 to 0.155) and adds only a small entropy reduction (0.514 to 0.499), as $r_f + r_a$ alone already lowers entropy from the baseline (0.615). Per-dataset entropy differences with 95\% confidence intervals are in Appendix Table~\ref{tab:entropy-ci}. The perception gain from ASM may come at the cost of symbolic reasoning; domain- and difficulty-aware reward balancing \citep{dai2026qoq, zhou2025disco} is a natural direction for future work.

\begin{table}[t]
\centering
\small
\setlength{\tabcolsep}{3pt}
\resizebox{\columnwidth}{!}{%
\begin{tabular}{lcccccccc}
\toprule
Method & Overall. & Phys. & Multi. & Spat. & Num. & Pred. & Impl. & Ent.$\downarrow$ \\
\midrule
\multicolumn{9}{l}{\textit{Molmo2-O-7B}} \\
Baseline & 0.057 & 0.055 & 0.030 & 0.078 & 0.104 & 0.039 & 0.086 & 0.740 \\
SFT & 0.056 & 0.052 & 0.041 & 0.073 & \textbf{0.125} & 0.015 & 0.052 & 0.736 \\
$r_f + r_a$ & 0.085 & 0.082 & 0.058 & 0.108 & 0.068 & \textbf{0.137} & 0.069 & 0.773 \\
$r_f + r_a$ + ASM & \textbf{0.097} & \textbf{0.093} & \textbf{0.062} & \textbf{0.135} & 0.083 & 0.108 & \textbf{0.109} & \textbf{0.683} \\
$\Delta$ & +0.041 & +0.039 & +0.033 & +0.057 & -0.021 & +0.069 & +0.023 & -0.057 \\
\midrule
\multicolumn{9}{l}{\textit{Gemma-4-E2B-it}} \\
Baseline & 0.047 & 0.042 & 0.029 & 0.052 & 0.141 & 0.020 & 0.011 & 0.615 \\
SFT & 0.049 & 0.042 & 0.048 & 0.051 & 0.089 & 0.020 & 0.034 & 0.573 \\
$r_f + r_a$ & \textbf{0.183} & \textbf{0.169} & \textbf{0.144} & \textbf{0.196} & \textbf{0.229} & \textbf{0.250} & 0.132 & 0.514 \\
$r_f + r_a$ + ASM & 0.171 & 0.162 & 0.140 & 0.173 & 0.193 & 0.225 & \textbf{0.155} & \textbf{0.499} \\
$\Delta$ & +0.124 & +0.120 & +0.111 & +0.121 & +0.052 & +0.206 & +0.144 & -0.116 \\
\bottomrule
\end{tabular}%
}
\caption{PhyX accuracy (judge score) by reasoning type for open-ended questions (top-3 run average), and normalized attention entropy (Ent.; lower = more concentrated) over all PhyX, MMMU, and SciBench test questions for the best checkpoint of each configuration. Reasoning types: Physical Model Grounding (Phys.), Multi-Formula (Multi.), Spatial Relation (Spat.), Numerical (Num.), Predictive (Pred.), and Implicit Condition (Impl.). $\Delta$ = last row minus baseline. Best per column and model in bold.}
\label{tab:reasoning-types}
\end{table}

\paragraph{Is the ASM gain due to attention grounding rather than the reward's structure?}

To test whether the effect of ASM comes from the model's own attention, we replace the attention map with a noise-perturbed version before scoring: Gaussian noise with standard deviation $\sigma \in \{0.1, 0.5, 1.0\}$ is added to the normalized patch map, drawn fresh for every completion (details in Appendix~\ref{app:noise_control}). The reward keeps ASM's foreground mask, score formula, and $[0,1]$ range, but no longer reflects where the model attends. In these runs the format reward gates the accuracy and ASM terms, so a noisy reward cannot pay out for malformed outputs. We train GRPO~($r_f + r_a$ + ASM) on Molmo~2 on open-ended questions under each noise level.

Perturbing the attention map consistently hurts (Table~\ref{tab:asm-controls-nostd}). Pooled accuracy drops from 0.115 (unperturbed ASM) to 0.066, 0.055, and 0.048 at $\sigma=0.1$, $0.5$, and $1.0$, decreasing monotonically with the noise level and on every benchmark. Every noisy variant is also well below $r_f + r_a$ (0.107 pooled under the same judge), so a reward that preserves ASM's scale and structure but carries no information about the model's attention is not a harmless regularizer: it degrades the policy, plausibly because the attention term injects reward variance that is uncorrelated with completion quality. This rules out the explanation that ASM's effect is only a consequence of its reward range or format gating, and shows that its benefit depends on the reward reflecting the model's actual attention.

\begin{table}[t]
\centering
\small
\setlength{\tabcolsep}{4pt}
\resizebox{\columnwidth}{!}{%
\begin{tabular}{lcccc}
\toprule
Reward & Avg & PhyX & MMMU & SciBench \\
\midrule
\multicolumn{5}{l}{\emph{Unperturbed}} \\
\quad $r_f + r_a$ + ASM & 0.115 & 0.097 & 0.155 & 0.142 \\
\midrule
\multicolumn{5}{l}{\emph{Attention noise ($\sigma$)}} \\
\quad $\sigma=0.1$ & 0.066 & 0.050 & 0.101 & 0.060 \\
\quad $\sigma=0.5$ & 0.055 & 0.039 & 0.086 & 0.055 \\
\quad $\sigma=1.0$ & 0.048 & 0.034 & 0.083 & 0.044 \\
\midrule
\multicolumn{5}{l}{\emph{Random mask}} \\
\quad Area-matched & 0.121 & 0.093 & 0.181 & 0.120 \\
\quad 10\% of patches & 0.115 & 0.089 & 0.173 & 0.107 \\
\quad 25\% of patches & 0.114 & 0.100 & 0.149 & 0.098 \\
\quad 50\% of patches & 0.115 & 0.099 & 0.156 & 0.148 \\
\bottomrule
\end{tabular}%
}
\caption{Controls for ASM on Molmo~2. All rows use $r_f + r_a$ + ASM and differ only in the ASM term's attention map or mask. Accuracy (judge score) is the mean over 3 runs. Noise: Gaussian noise with std $\sigma$ added to the normalized attention map. Random mask: a random mask replaces the foreground mask, with the same area (area-matched) or a fixed fraction of patches.}
\label{tab:asm-controls-nostd}
\end{table}

\paragraph{Does ASM reward visual grounding or just attention to non-white pixels?}
A concern with ASM is that the foreground mask (non-white pixels) does not distinguish physically relevant content from other non-white pixels: a model could raise its score by attending to labels, arrows, or option text rather than the regions that matter for the problem. To test whether the mask carries useful signal, we replace it with a \textit{random binary mask} on the same patch grid, keeping the ASM computation, reward range, and the model's real attention weights unchanged, so only the correspondence between the mask and the image content is destroyed. The main variant matches the total area of the true foreground mask; we also run fixed-fraction masks covering $p \in \{10\%, 25\%, 50\%\}$ of the patches (details in Appendix~\ref{app:randmask_control}). As in the noise control, the format reward gates the other reward terms, and we train GRPO~($r_f + r_a$ + ASM) on Molmo~2 on open-ended questions. If the true mask outperforms the random masks, particularly on Spatial Relation, MMMU, and SciBench, it provides a grounding prior beyond the reward's scale and structure; if they perform comparably, the gain is better explained by regularization than by the mask's visual grounding.

The random-mask rewards perform comparably to the true-mask reward (Table~\ref{tab:asm-controls-nostd}). Pooled accuracy is 0.114--0.121 across the four random-mask variants versus 0.115 for the true mask, and the area-matched mask is nominally higher, in particular on MMMU (0.181 vs.\ 0.155). PhyX accuracy spans 0.089--0.100 (true mask 0.097), and the variants differ most on SciBench (0.098--0.148 vs.\ 0.142), the smallest benchmark, where differences of this size are within the run-to-run variability. The true mask is higher than the area-matched and 10\% masks on PhyX Spatial Relation reasoning (0.135 vs.\ 0.102 and 0.105), but the 25\% and 50\% masks reach 0.135 and 0.143 (one run each), so we cannot attribute the spatial gain to the content of the foreground mask. Together with the noise control, these results suggest that ASM's effect depends on the reward being computed from the model's own attention, but not on where the mask lies: random regions of the same or a different size yield similar accuracy. 

\begin{figure*}[t]
    \centering
    \begin{subfigure}[b]{0.66\textwidth}
        \centering
        \includegraphics[width=\linewidth]{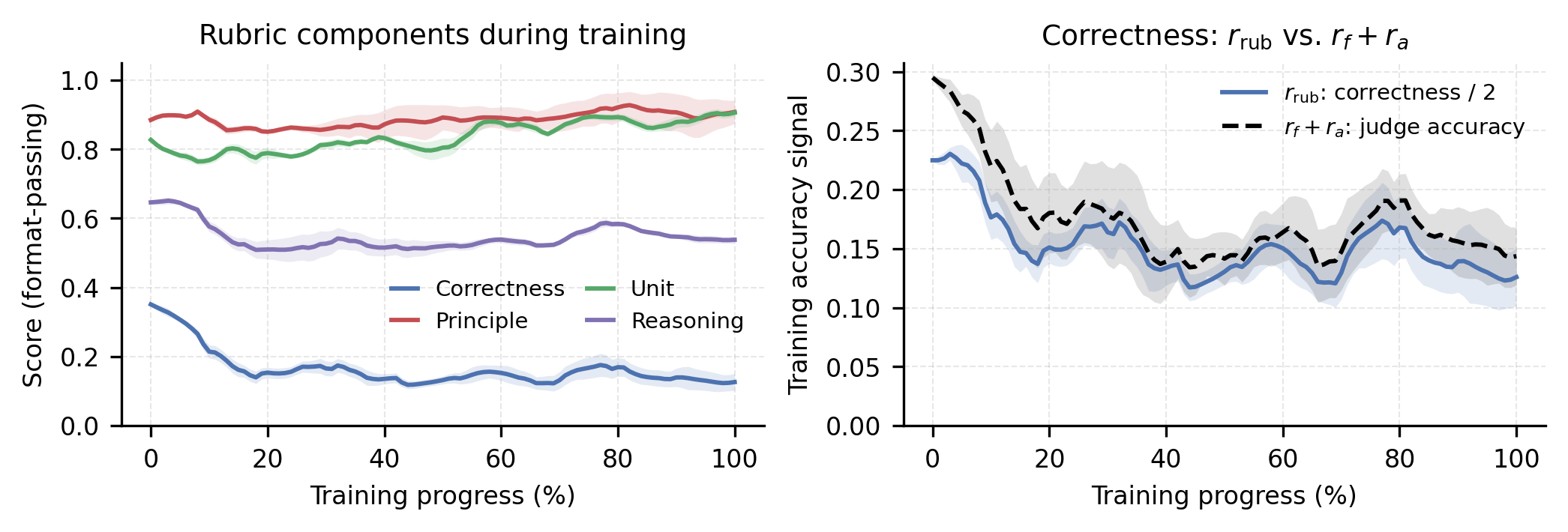}
        \caption{Training dynamics.}
        \label{fig:rubric_curves}
    \end{subfigure}
    \hfill
    \begin{subfigure}[b]{0.31\textwidth}
        \centering
        \includegraphics[width=\linewidth]{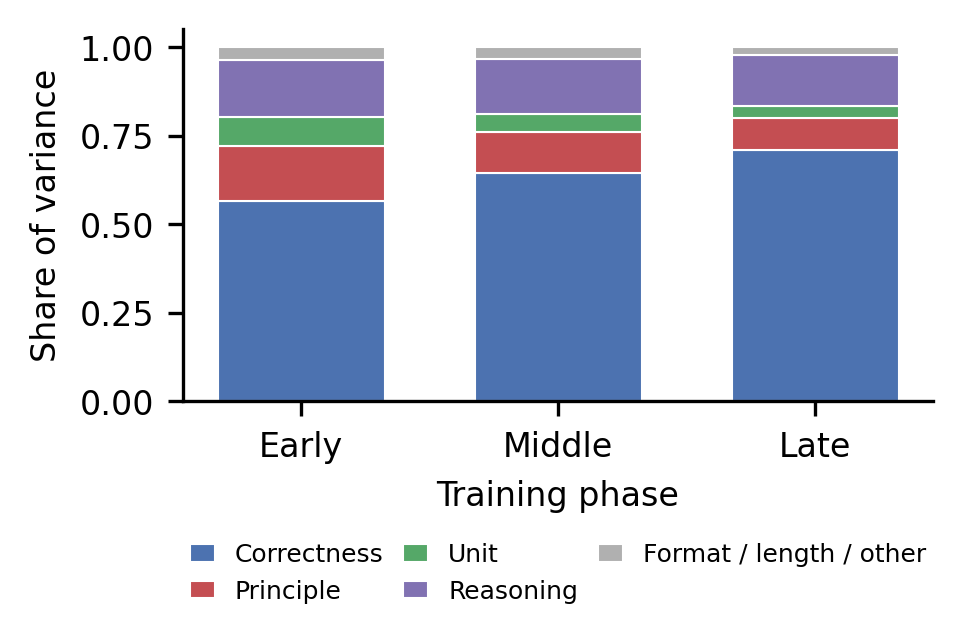}
        \caption{Reward variance share.}
        \label{fig:rubric_share}
    \end{subfigure}
    \caption{Rubric reward analysis for Gemma~4 (Molmo~2 in Appendix Figure~\ref{fig:rubric_molmo2}). (a) Mean $\pm$ std over seeds (7-step moving average) against the percentage of training. Left: component scores of format-passing completions, normalized to $[0,1]$. Right: rubric correctness (divided by 2) over all completions. (b) Share of the within-group reward variance attributable to each rubric term in the early, middle, and late thirds of training.}
    \label{fig:rubric}
\end{figure*}

\paragraph{Does the rubric's principle and unit components contribute beyond accuracy alone?}

The rubric reward $r_{\text{rub}}$ scores each completion on four judge-graded dimensions from a single judge call: correctness (0--2), principle (0/1), unit (0/1), and reasoning (0--2). The training logs therefore let us ask how much each component contributes to the learning signal, without extra training. We analyze every rollout logged while training the rubric reward on Molmo~2 (6 seeds) and Gemma~4 (3 seeds), reconstructing each GRPO group (8 completions of one prompt) from the per-rank logs, and compare with the $r_f + r_a$ runs (weights and reconstruction details in Appendix~\ref{app:rubric_details}). As Section 4 shows, the rubric reward does not exceed $r_f + r_a$ in accuracy on either model (pooled 0.058 vs.\ 0.107 on Molmo~2 and 0.200 vs.\ 0.209 on Gemma~4; Table~\ref{tab:oe-sft-avg}), so the question is what its extra components add.

\emph{Training improves the secondary components, not correctness.} Figure~\ref{fig:rubric_molmo2} (left) shows that on Molmo~2 the principle, unit, and reasoning scores of format-passing completions rise steadily (first vs.\ last 10 steps: principle $0.80\!\to\!0.90$, unit $0.70\!\to\!0.86$, reasoning $0.38\!\to\!0.50$), while correctness stays near $0.06$ (of a maximum of 1). On Gemma~4 (Figure~\ref{fig:rubric_curves}, left), principle and unit are already high ($0.88\!\to\!0.90$, $0.77\!\to\!0.88$), reasoning decreases slightly ($0.60\!\to\!0.55$), and correctness falls from $0.26$ to $0.14$. The rubric's correctness term tracks the $r_f + r_a$ judge accuracy on Molmo~2 and is lower early in training on Gemma~4 (Figures~\ref{fig:rubric_curves} and~\ref{fig:rubric_molmo2}, right; the two use different judge prompts, so we compare only trends). Adding principle, unit, and reasoning therefore does not translate into higher correctness during training.

\emph{Correctness is sparse, so secondary terms often supply the learning signal.} Only 7\% (Molmo~2) and 17\% (Gemma~4) of format-passing training completions are judged correct (correctness $\geq 1$ of 2), so correctness is identical across all completions of a group in 60--78\% of groups and contributes no advantage there. A correctness-only reward would give a learning signal in 28\% (Molmo~2) and 39\% (Gemma~4) of groups, versus 89\% and 95\% for the full rubric (Appendix Table~\ref{tab:rubric-reweight}). Correctness still accounts for the largest share of within-group reward variance (52\% and 63\%; Figures~\ref{fig:rubric_share} and~\ref{fig:rubric_molmo2}), but principle and unit together account for 28\% (Molmo~2) and 19\% (Gemma~4) and reasoning for a further 16\% and 15\%. The advantages implied by the full rubric are close to those of its secondary terms alone (correlation 0.93 and 0.92 for principle+unit+reasoning, vs.\ 0.89 and 0.94 for correctness only), so a substantial part of the update direction is not driven by correctness.

\emph{Principle and unit are weak proxies for correctness; reasoning is informative.} Within a group, principle and unit correlate only weakly with correctness ($r=0.26$ and $0.24$ on Molmo~2, $0.42$ and $0.35$ on Gemma~4), whereas reasoning correlates strongly ($r=0.64$ and $0.72$; Appendix Table~\ref{tab:rubric-validity}). Principle and unit are also satisfied by the large majority of completions ($>$80\% late in training), so they discriminate little. Since one judge call produces all components, part of the correlation reflects judge consistency, and these values are an upper bound on independent information.

Taken together, the principle and unit terms do not contribute beyond accuracy in our setting: they are easy to satisfy, weakly related to correctness, and optimized without a gain in correctness. Reasoning is the one component that carries signal related to correctness, but adding it does not improve accuracy over $r_f + r_a$ either.

%% file: sections/6_Conclusion.tex
We studied how external rewards (format, answer accuracy, rubric) and an internal attention-mask reward (ASM) shape physical reasoning in two vision-language models trained with GRPO on open-ended PhyX questions. External answer-accuracy supervision drives most of the accuracy gains, in-domain and zero-shot on MMMU and SciBench, while SFT and format-only rewards give little or no gain. ASM complements it: it needs no labels or judge, yet on Molmo~2 adding it to $r_f + r_a$ gives the best PhyX accuracy, improves the visually demanding Spatial Relation reasoning, concentrates attention on image content, and preserves out-of-distribution accuracy, and on Gemma~4 it stays within one standard deviation of $r_f + r_a$. Control experiments show that this benefit depends on the reward reflecting the model's own attention, and identifying exactly what ASM supervises is a promising direction for future work. The rubric's principle and unit components, in contrast, contribute little beyond accuracy.

Overall, our results suggest that internal attention rewards are a low-cost way to supplement external supervision, and that the reward should be matched to what matters in deployment, whether raw accuracy, interpretable reasoning, or visual grounding.

\section*{Limitations}
It covers two models, open-ended questions only, and one in-domain benchmark (PhyX) plus zero-shot MMMU and SciBench evaluation. Each configuration is trained three times (fewer for some control variants) and several show large run-to-run variability, so small differences should be interpreted with caution. The attention-based reward is examined with limited controls and only on one model, and the accuracy and rubric rewards rely on a single LLM judge. Whether our findings generalize to other model families, larger VLMs, and multiple-choice formats remains an open question for future work.

\section*{Potential Risks}
Our work trains vision-language models on physics problems with public benchmarks and no personal data, so we see little direct risk. Models trained with our rewards can still produce confident but incorrect physics solutions, and should not be relied on for education or engineering decisions without verification. Rewards that depend on LLM judges can be exploited by the policy or inherit judge errors and biases, and the attention-based reward may encourage attending to non-white image content without improving reasoning, so its attention maps should not be read as a faithful explanation of the model's reasoning. GRPO training also requires substantial GPU compute.

\section*{Use of AI Assistants}
We used AI assistants for some of the writing (drafting and editing text and tables) and for some of the coding. The authors reviewed all AI-assisted content and take full responsibility for the paper.

\section*{Acknowledgment}

This work was supported by the U.S. Department of Energy, Advanced Scientific Computing Research program and Pacific Northwest National Laboratory (PNNL), which is operated by Battelle Memorial Institute for the U.S. Department of Energy under Contract DEAC05–76RLO1830. This research used resources of the National Energy Research Scientific Computing Center (NERSC), a Department of Energy User Facility using NERSC award ASCR-ERCAP0038273. This work is approved for release by PNNL under IR: PNNL-SA-228365.

%% file: sections/7_Appendix.tex
\raggedbottom
\section{Additional Results}

\subsection{Open-Ended Results with Run-to-Run Variability}

Table~\ref{tab:oe-sft-avg-std} reports the open-ended results of Table~\ref{tab:oe-sft-avg} together with the sample standard deviation over the top-3 runs of each configuration, for both models. Standard deviations are computed per column over the runs that make up each mean. Variability is substantial for several GRPO configurations (e.g., Gemma~4 with $r_f$ and $r_f$ + ASM, and the Molmo2 and Gemma~4 rubric reward $r_{\text{rub}}$), so differences smaller than one standard deviation should not be interpreted as a consistent ranking of reward configurations.

\begin{table*}[!tb]
\centering
\small
\resizebox{\ifdim\width>\textwidth\textwidth\else\width\fi}{!}{%
\begin{tabular}{lccccccccccccccc}
\toprule
 &  & \multicolumn{7}{c}{PhyX} & \multicolumn{6}{c}{MMMU} & \multicolumn{1}{c}{SciBench} \\
\cmidrule(lr){3-9}\cmidrule(lr){10-15}\cmidrule(lr){16-16}
Method & Avg & P-Ovr & P-EM & P-Mech & P-Mod & P-Opt & P-Th & P-W/A & M-Ovr & M-CM & M-EM & M-NP & M-Opt & M-TSM & S-Fund \\
\midrule
\multicolumn{1}{c}{\textbf{Gemma-4-E2B-it}} \\
Baseline & 0.066 {\scriptsize $\pm$0.004} & 0.047 {\scriptsize $\pm$0.002} & 0.077 {\scriptsize $\pm$0.006} & 0.073 {\scriptsize $\pm$0.003} & 0.049 {\scriptsize $\pm$0.000} & 0.023 {\scriptsize $\pm$0.010} & 0.023 {\scriptsize $\pm$0.007} & 0.036 {\scriptsize $\pm$0.006} & 0.110 {\scriptsize $\pm$0.010} & 0.104 {\scriptsize $\pm$0.009} & 0.096 {\scriptsize $\pm$0.004} & 0.167 {\scriptsize $\pm$0.056} & 0.187 {\scriptsize $\pm$0.031} & 0.057 {\scriptsize $\pm$0.029} & 0.060 {\scriptsize $\pm$0.009} \\
SFT & 0.057 {\scriptsize $\pm$0.004} & 0.049 {\scriptsize $\pm$0.004} & 0.069 {\scriptsize $\pm$0.007} & 0.062 {\scriptsize $\pm$0.015} & 0.045 {\scriptsize $\pm$0.009} & 0.036 {\scriptsize $\pm$0.010} & 0.041 {\scriptsize $\pm$0.004} & 0.040 {\scriptsize $\pm$0.009} & 0.072 {\scriptsize $\pm$0.011} & 0.069 {\scriptsize $\pm$0.009} & 0.073 {\scriptsize $\pm$0.008} & 0.130 {\scriptsize $\pm$0.032} & 0.053 {\scriptsize $\pm$0.050} & 0.076 {\scriptsize $\pm$0.016} & 0.044 {\scriptsize $\pm$0.019} \\
GRPO ($r_f$) & 0.101 {\scriptsize $\pm$0.045} & 0.071 {\scriptsize $\pm$0.041} & 0.067 {\scriptsize $\pm$0.033} & 0.103 {\scriptsize $\pm$0.060} & 0.065 {\scriptsize $\pm$0.049} & 0.046 {\scriptsize $\pm$0.026} & 0.060 {\scriptsize $\pm$0.046} & 0.085 {\scriptsize $\pm$0.034} & 0.152 {\scriptsize $\pm$0.044} & 0.143 {\scriptsize $\pm$0.034} & 0.136 {\scriptsize $\pm$0.055} & 0.167 {\scriptsize $\pm$0.000} & 0.200 {\scriptsize $\pm$0.072} & 0.181 {\scriptsize $\pm$0.072} & 0.082 {\scriptsize $\pm$0.043} \\
GRPO ($r_f$ + ASM) & 0.080 {\scriptsize $\pm$0.055} & 0.053 {\scriptsize $\pm$0.044} & 0.048 {\scriptsize $\pm$0.046} & 0.079 {\scriptsize $\pm$0.068} & 0.049 {\scriptsize $\pm$0.043} & 0.038 {\scriptsize $\pm$0.055} & 0.037 {\scriptsize $\pm$0.016} & 0.067 {\scriptsize $\pm$0.042} & 0.127 {\scriptsize $\pm$0.080} & 0.119 {\scriptsize $\pm$0.059} & 0.108 {\scriptsize $\pm$0.092} & 0.185 {\scriptsize $\pm$0.085} & 0.167 {\scriptsize $\pm$0.110} & 0.162 {\scriptsize $\pm$0.100} & 0.060 {\scriptsize $\pm$0.034} \\
GRPO ($r_f + r_a$) & 0.209 {\scriptsize $\pm$0.025} & 0.183 {\scriptsize $\pm$0.033} & 0.131 {\scriptsize $\pm$0.016} & 0.284 {\scriptsize $\pm$0.063} & 0.203 {\scriptsize $\pm$0.051} & 0.137 {\scriptsize $\pm$0.041} & 0.133 {\scriptsize $\pm$0.040} & 0.204 {\scriptsize $\pm$0.039} & 0.255 {\scriptsize $\pm$0.009} & 0.268 {\scriptsize $\pm$0.034} & 0.202 {\scriptsize $\pm$0.016} & 0.444 {\scriptsize $\pm$0.096} & 0.287 {\scriptsize $\pm$0.012} & 0.257 {\scriptsize $\pm$0.029} & 0.213 {\scriptsize $\pm$0.028} \\
GRPO ($r_f + r_a$ + ASM) & 0.201 {\scriptsize $\pm$0.014} & 0.171 {\scriptsize $\pm$0.012} & 0.127 {\scriptsize $\pm$0.040} & 0.268 {\scriptsize $\pm$0.021} & 0.169 {\scriptsize $\pm$0.025} & 0.124 {\scriptsize $\pm$0.049} & 0.143 {\scriptsize $\pm$0.012} & 0.192 {\scriptsize $\pm$0.013} & 0.251 {\scriptsize $\pm$0.019} & 0.259 {\scriptsize $\pm$0.036} & 0.221 {\scriptsize $\pm$0.025} & 0.352 {\scriptsize $\pm$0.179} & 0.240 {\scriptsize $\pm$0.040} & 0.295 {\scriptsize $\pm$0.059} & 0.191 {\scriptsize $\pm$0.009} \\
GRPO ($r_{\text{rub}}$) & 0.200 {\scriptsize $\pm$0.026} & 0.178 {\scriptsize $\pm$0.025} & 0.125 {\scriptsize $\pm$0.012} & 0.278 {\scriptsize $\pm$0.042} & 0.193 {\scriptsize $\pm$0.040} & 0.133 {\scriptsize $\pm$0.019} & 0.112 {\scriptsize $\pm$0.028} & 0.220 {\scriptsize $\pm$0.024} & 0.236 {\scriptsize $\pm$0.045} & 0.223 {\scriptsize $\pm$0.055} & 0.204 {\scriptsize $\pm$0.037} & 0.407 {\scriptsize $\pm$0.085} & 0.247 {\scriptsize $\pm$0.012} & 0.324 {\scriptsize $\pm$0.092} & 0.219 {\scriptsize $\pm$0.025} \\
\midrule
\multicolumn{1}{c}{\textbf{Molmo2-O-7B}} \\
Baseline & 0.082 {\scriptsize $\pm$0.006} & 0.057 {\scriptsize $\pm$0.005} & 0.062 {\scriptsize $\pm$0.009} & 0.077 {\scriptsize $\pm$0.006} & 0.051 {\scriptsize $\pm$0.009} & 0.038 {\scriptsize $\pm$0.000} & 0.046 {\scriptsize $\pm$0.013} & 0.065 {\scriptsize $\pm$0.014} & 0.144 {\scriptsize $\pm$0.006} & 0.160 {\scriptsize $\pm$0.009} & 0.122 {\scriptsize $\pm$0.020} & 0.352 {\scriptsize $\pm$0.064} & 0.100 {\scriptsize $\pm$0.035} & 0.105 {\scriptsize $\pm$0.033} & 0.022 {\scriptsize $\pm$0.019} \\
SFT & 0.077 {\scriptsize $\pm$0.001} & 0.056 {\scriptsize $\pm$0.002} & 0.071 {\scriptsize $\pm$0.000} & 0.063 {\scriptsize $\pm$0.003} & 0.063 {\scriptsize $\pm$0.004} & 0.034 {\scriptsize $\pm$0.004} & 0.035 {\scriptsize $\pm$0.004} & 0.065 {\scriptsize $\pm$0.003} & 0.125 {\scriptsize $\pm$0.002} & 0.134 {\scriptsize $\pm$0.000} & 0.113 {\scriptsize $\pm$0.000} & 0.278 {\scriptsize $\pm$0.056} & 0.120 {\scriptsize $\pm$0.000} & 0.057 {\scriptsize $\pm$0.000} & 0.016 {\scriptsize $\pm$0.000} \\
GRPO ($r_f$) & 0.061 {\scriptsize $\pm$0.005} & 0.045 {\scriptsize $\pm$0.007} & 0.058 {\scriptsize $\pm$0.003} & 0.054 {\scriptsize $\pm$0.016} & 0.030 {\scriptsize $\pm$0.006} & 0.030 {\scriptsize $\pm$0.010} & 0.031 {\scriptsize $\pm$0.011} & 0.067 {\scriptsize $\pm$0.022} & 0.087 {\scriptsize $\pm$0.012} & 0.089 {\scriptsize $\pm$0.010} & 0.073 {\scriptsize $\pm$0.029} & 0.222 {\scriptsize $\pm$0.000} & 0.080 {\scriptsize $\pm$0.035} & 0.076 {\scriptsize $\pm$0.016} & 0.093 {\scriptsize $\pm$0.009} \\
GRPO ($r_f$ + ASM) & 0.062 {\scriptsize $\pm$0.015} & 0.045 {\scriptsize $\pm$0.012} & 0.050 {\scriptsize $\pm$0.003} & 0.052 {\scriptsize $\pm$0.019} & 0.049 {\scriptsize $\pm$0.016} & 0.023 {\scriptsize $\pm$0.010} & 0.052 {\scriptsize $\pm$0.031} & 0.046 {\scriptsize $\pm$0.009} & 0.098 {\scriptsize $\pm$0.018} & 0.101 {\scriptsize $\pm$0.040} & 0.077 {\scriptsize $\pm$0.000} & 0.204 {\scriptsize $\pm$0.032} & 0.093 {\scriptsize $\pm$0.031} & 0.124 {\scriptsize $\pm$0.044} & 0.060 {\scriptsize $\pm$0.034} \\
GRPO ($r_f + r_a$) & 0.107 {\scriptsize $\pm$0.023} & 0.085 {\scriptsize $\pm$0.021} & 0.058 {\scriptsize $\pm$0.009} & 0.131 {\scriptsize $\pm$0.033} & 0.065 {\scriptsize $\pm$0.031} & 0.061 {\scriptsize $\pm$0.018} & 0.066 {\scriptsize $\pm$0.013} & 0.129 {\scriptsize $\pm$0.039} & 0.147 {\scriptsize $\pm$0.018} & 0.145 {\scriptsize $\pm$0.006} & 0.138 {\scriptsize $\pm$0.028} & 0.222 {\scriptsize $\pm$0.056} & 0.147 {\scriptsize $\pm$0.099} & 0.152 {\scriptsize $\pm$0.044} & 0.148 {\scriptsize $\pm$0.057} \\
GRPO ($r_f + r_a$ + ASM) & 0.115 {\scriptsize $\pm$0.013} & 0.097 {\scriptsize $\pm$0.013} & 0.075 {\scriptsize $\pm$0.028} & 0.139 {\scriptsize $\pm$0.038} & 0.073 {\scriptsize $\pm$0.018} & 0.084 {\scriptsize $\pm$0.026} & 0.077 {\scriptsize $\pm$0.018} & 0.133 {\scriptsize $\pm$0.024} & 0.155 {\scriptsize $\pm$0.024} & 0.151 {\scriptsize $\pm$0.022} & 0.143 {\scriptsize $\pm$0.008} & 0.185 {\scriptsize $\pm$0.085} & 0.133 {\scriptsize $\pm$0.031} & 0.238 {\scriptsize $\pm$0.072} & 0.142 {\scriptsize $\pm$0.038} \\
GRPO ($r_{\text{rub}}$) & 0.058 {\scriptsize $\pm$0.051} & 0.049 {\scriptsize $\pm$0.043} & 0.014 {\scriptsize $\pm$0.015} & 0.091 {\scriptsize $\pm$0.084} & 0.037 {\scriptsize $\pm$0.032} & 0.038 {\scriptsize $\pm$0.033} & 0.041 {\scriptsize $\pm$0.036} & 0.073 {\scriptsize $\pm$0.064} & 0.073 {\scriptsize $\pm$0.064} & 0.067 {\scriptsize $\pm$0.058} & 0.070 {\scriptsize $\pm$0.061} & 0.019 {\scriptsize $\pm$0.032} & 0.113 {\scriptsize $\pm$0.099} & 0.086 {\scriptsize $\pm$0.076} & 0.060 {\scriptsize $\pm$0.053} \\
\bottomrule
\end{tabular}%
}
\caption{Mean $\pm$ sample standard deviation of the judge score over the top-3 runs of each configuration on PhyX (P), MMMU (M) and SciBench (S); a dash after $\pm$ means only one run. GRPO rows are labeled by their reward terms: $r_f$ = rule-based format reward, $r_a$ = LLM-judge (jury) answer accuracy, $r_{\text{rub}}$ = LLM-judge (rubric) answer and physics reasoning accuracy, ASM = attention-mask score. Column labels are \texttt{<dataset>-<category>}. \textbf{P-} = PhyX: Overall (Ovr), Electromagnetism (EM), Mechanics (Mech), Modern Physics (Mod), Optics (Opt), Thermodynamics (Th), Waves and Acoustics (W/A). \textbf{M-} = MMMU: Overall (Ovr), Classical Mechanics (CM), Electromagnetism (EM), Nuclear Physics (NP), Optics (Opt), Thermodynamics and Statistical Mechanics (TSM). \textbf{S-} = SciBench: Fundamentals of Physics (Fund). Avg = sample-weighted average of the datasets' overall scores.}
\label{tab:oe-sft-avg-std}
\end{table*}

\section{Dataset Statistics}
\label{appendix:dataset_statistics}
This section describes the labels used by the rubric's principle and unit terms for the open-ended PhyX data used throughout (2,016 training and 984 test questions; the test split is the PhyX testmini subset). The numbers below were recomputed from the released data files and replace the earlier statistics. Table~\ref{tab:phyx_dataset_stats} gives the physics domains.

\subsection{Principle Labels}
Each question carries the PhyX subfield (Table~\ref{tab:phyx_subfield}; 26 subfields). We do not train against a fixed principle string: during training the rubric judge sees the subfield as a coarse hint about the area of physics and decides whether the stated principle correctly governs the problem. The distribution is long-tailed: the five most frequent subfields (Geometrical Optics, Laws of Thermodynamics, Wave Properties, Electrostatics, Quantum Phenomena) cover 46\% of the questions, and 15 of the 26 subfields have fewer than 100 questions.

\subsection{Unit Labels}
Each question has a ground-truth unit, a physical quantity category, and an SI prefix. The unit labels are created in two steps. First, a rule-based parser (\texttt{extract\_units.py}) normalizes the unit notation in the answer and assigns a label only when the unit is unambiguous; this labels 2,378 of the 3,000 questions (79\%). Second, the remaining 622 questions are labeled by GPT-5, which sees the question, answer, and image and must choose a quantity category from those found by the parser, a canonical unit name, and an SI prefix, and flags the label when the evidence is ambiguous (the prompt is given in Appendix~\ref{app:system_prompt_unit_labeling}); 243 labels were flagged. Seven questions have no unit label (symbolic or qualitative answers). The labels cover 71 quantity categories and 135 distinct units (Tables~\ref{tab:clusters} and~\ref{tab:unit_strings}). Length, energy, velocity, angle, and force account for 56\% of the questions, and 78\% of the questions with a prefix label are unprefixed (kilo and centi are the most common prefixes, at 7\% and 6\%). Unit labels are shared between the multiple-choice and open-ended versions of a question.

\begin{table*}[!tb]
\centering
\small
\resizebox{\ifdim\width>\textwidth\textwidth\else\width\fi}{!}{%
\begin{tabular}{lrrr}
\toprule
\textbf{Domain} & Train & Test & Total \\
\midrule
Electromagnetism & 382 & 168 & 550 \\
Mechanics & 382 & 168 & 550 \\
Optics & 342 & 158 & 500 \\
Thermodynamics & 339 & 161 & 500 \\
Waves/Acoustics & 335 & 165 & 500 \\
Modern Physics & 236 & 164 & 400 \\
\midrule
\multicolumn{4}{l}{Total: 2016 train, 984 test, 3000 overall} \\
\bottomrule
\end{tabular}%
}
\caption{PhyX open-ended questions per physics domain in the train and test splits.}
\label{tab:phyx_dataset_stats}
\end{table*}

\begin{table*}[!tb]
\centering
\small
\resizebox{\ifdim\width>\textwidth\textwidth\else\width\fi}{!}{%
\begin{tabular}{lrrr@{\hspace{1.2em}}lrrr}
\toprule
\textbf{Principle (subfield)} & Train & Test & Total & \textbf{Principle (subfield)} & Train & Test & Total \\
\midrule
Geometrical Optics & 270 & 127 & 397 & Statics & 57 & 21 & 78 \\
Laws of Thermodynamics & 189 & 90 & 279 & Magnetism & 50 & 25 & 75 \\
Wave Properties & 186 & 87 & 273 & Rotational Motion & 33 & 39 & 72 \\
Electrostatics & 145 & 81 & 226 & Sound & 54 & 18 & 72 \\
Quantum Phenomena & 100 & 101 & 201 & Ideal Gases and Kinetic Theory & 33 & 24 & 57 \\
Dynamics & 154 & 34 & 188 & Momentum and Collisions & 20 & 9 & 29 \\
Relativity & 115 & 47 & 162 & Nuclear Physics & 14 & 10 & 24 \\
Electric Circuits & 121 & 24 & 145 & Electromagnetic Waves & 12 & 4 & 16 \\
Temperature and Heat Transfer & 102 & 35 & 137 & Optical Instruments & 11 & 4 & 15 \\
Resonance and Harmonics & 77 & 52 & 129 & Specific Heat and Calorimetry & 7 & 5 & 12 \\
Kinematics & 73 & 54 & 127 & Particle Physics & 5 & 4 & 9 \\
Wave Optics & 74 & 30 & 104 & Geometry & 0 & 1 & 1 \\
Electromagnetic Induction & 52 & 36 & 88 & Fluid Mechanics & 1 & 0 & 1 \\
Work and Energy & 61 & 22 & 83 &  &  &  &  \\
\midrule
\multicolumn{8}{l}{Total: 2016 train, 984 test, 3000 overall} \\
\bottomrule
\end{tabular}%
}
\caption{Principle labels: the PhyX subfield of each question, which the rubric judge receives as a coarse hint about the area of physics (it accepts any principle that correctly governs the problem). Counts of open-ended questions in the train and test splits (27 labels).}
\label{tab:phyx_subfield}
\end{table*}

\begin{table*}[!tb]
\centering
\small
\resizebox{\ifdim\width>\textwidth\textwidth\else\width\fi}{!}{%
\begin{tabular}{lrrr@{\hspace{1.2em}}lrrr}
\toprule
\textbf{Quantity category} & Train & Test & Total & \textbf{Quantity category} & Train & Test & Total \\
\midrule
Length & 441 & 211 & 652 & Electric Current & 40 & 18 & 58 \\
Energy & 217 & 106 & 323 & Pressure & 26 & 26 & 52 \\
Velocity & 212 & 104 & 316 & Mass & 33 & 17 & 50 \\
Angle & 123 & 82 & 205 & Acceleration & 36 & 11 & 47 \\
Force & 149 & 48 & 197 & Magnetic Field (Flux Density) & 26 & 18 & 44 \\
Time & 119 & 51 & 170 & Electric Charge & 20 & 12 & 32 \\
Frequency & 81 & 40 & 121 & Angular Velocity & 15 & 16 & 31 \\
Electric Potential & 67 & 21 & 88 & Electric Resistance & 24 & 6 & 30 \\
Power & 62 & 17 & 79 & Photon Count & 21 & 8 & 29 \\
Temperature & 46 & 20 & 66 & Other (51 categories) & 221 & 129 & 350 \\
Electric Field & 37 & 23 & 60 &  &  &  &  \\
\midrule
\multicolumn{8}{l}{Total: 2016 train, 984 test, 3000 overall} \\
\bottomrule
\end{tabular}%
}
\caption{Unit labels grouped by physical quantity category (71 categories in total; the 20 most frequent are listed). Counts of open-ended questions in the train and test splits.}
\label{tab:clusters}
\end{table*}

\begin{table*}[!tb]
\centering
\small
\resizebox{\ifdim\width>\textwidth\textwidth\else\width\fi}{!}{%
\begin{tabular}{lrrr@{\hspace{1.2em}}lrrr}
\toprule
\textbf{Unit} & Train & Test & Total & \textbf{Unit} & Train & Test & Total \\
\midrule
meter & 428 & 202 & 630 & hertz & 79 & 40 & 119 \\
dimensionless & 152 & 99 & 251 & volt & 66 & 21 & 87 \\
meter per second & 158 & 69 & 227 & watt & 55 & 12 & 67 \\
joule & 157 & 66 & 223 & ampere & 40 & 17 & 57 \\
newton & 147 & 46 & 193 & electronvolt & 38 & 18 & 56 \\
second & 102 & 48 & 150 & Other (123 units) & 501 & 291 & 792 \\
degree & 89 & 52 & 141 &  &  &  &  \\
\midrule
\multicolumn{8}{l}{Total: 2012 train, 981 test, 2993 overall} \\
\bottomrule
\end{tabular}%
}
\caption{Most frequent ground-truth unit strings supplied to the rubric judge (135 distinct units; 7 questions have no unit label because the answer is symbolic or qualitative).}
\label{tab:unit_strings}
\end{table*}

\subsection{Image Whitespace}
\label{appendix:whitespace}
\begin{figure}[!htbp]
\centering
\includegraphics[width=0.7\linewidth]{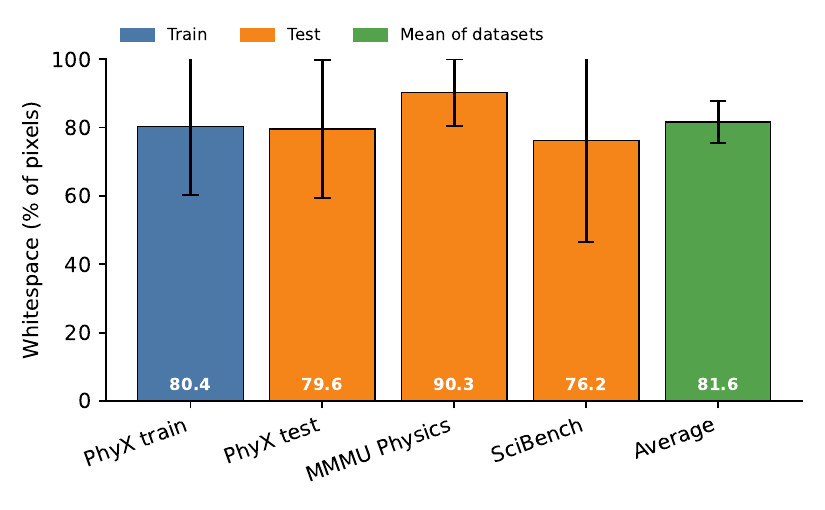}
\caption{Mean whitespace percentage per dataset. Error bars show the standard deviation over images (clipped at 100\%); the last bar is the mean of the four dataset means.}
\label{fig:whitespace_stats}
\end{figure}

The ASM reward (Equation~\ref{eq:r_attn}) is computed only over non-white pixels, so the amount of whitespace in an image determines how small the foreground region is. We measure it with the same rule as the reward: a pixel is whitespace if all its RGB channels are $\geq 230$. Table~\ref{tab:whitespace-stats} and Figure~\ref{fig:whitespace_stats} report the percentage of whitespace per image, summarized as mean $\pm$ standard deviation over the images of each dataset. On average $81.6\%$ of the pixels are whitespace ($81.1\%$ pooled over all 3,601 images), with MMMU Physics the sparsest ($90.3\%$) and SciBench the most variable. The PhyX training and test images have nearly the same whitespace share ($80.4\%$ vs.\ $79.6\%$), so the foreground area the reward sees during training matches the PhyX test set. If enclosed white regions are also counted as foreground (the hole-filling option of the mask), the average falls to $67.2\%$.

\begin{table*}[!tb]
\centering
\small
\resizebox{\ifdim\width>\textwidth\textwidth\else\width\fi}{!}{%
\begin{tabular}{llrr}
\toprule
Dataset & Split & \# images & Whitespace (\%) \\
\midrule
PhyX & Train & 2,016 & $80.4 \pm 20.3$ \\
PhyX & Test & 984 & $79.6 \pm 20.2$ \\
MMMU Physics & Test & 424 & $90.3 \pm 9.9$ \\
SciBench & Test & 177 & $76.2 \pm 29.6$ \\
\midrule
Average over datasets & & 3,601 & $81.6$ \\
Average over all images & & & $81.1 \pm 20.2$ \\
\bottomrule
\end{tabular}%
}
\caption{Percentage of image pixels classified as whitespace by the ASM reward (all RGB channels $\geq 230$), mean $\pm$ standard deviation over images.}
\label{tab:whitespace-stats}
\end{table*}

\section{Additional Implementation Details}

\subsection{Attention Entropy Score}
\label{appendix:attention_entropy_score}
\begin{figure}[!htbp]
    \centering
    \includegraphics[width=1\linewidth]{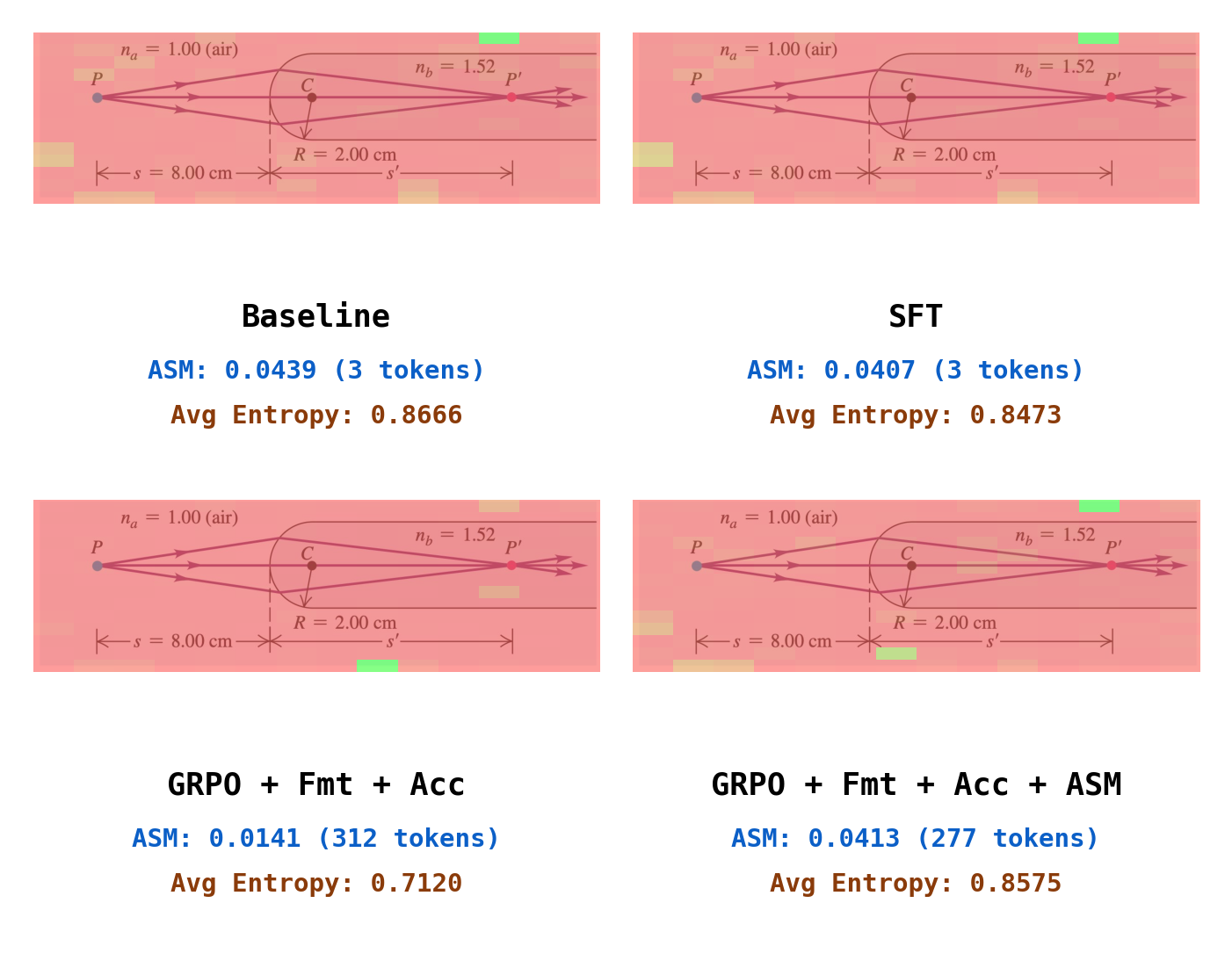}
    \caption{Cumulative image attention over the generated tokens for Molmo2-O-7B on a PhyX optics question (refraction at a spherical surface), for the baseline, SFT, GRPO ($r_f + r_a$), and GRPO ($r_f + r_a$ + ASM). Red indicates lower and green higher attention. ASM is the mean per-token attention-mask score and Avg Entropy is the normalized entropy of the cumulative map.}
    \label{fig:molmo2_phyx_attention}
\end{figure}

\begin{figure}[!htbp]
    \centering
    \includegraphics[width=1\linewidth]{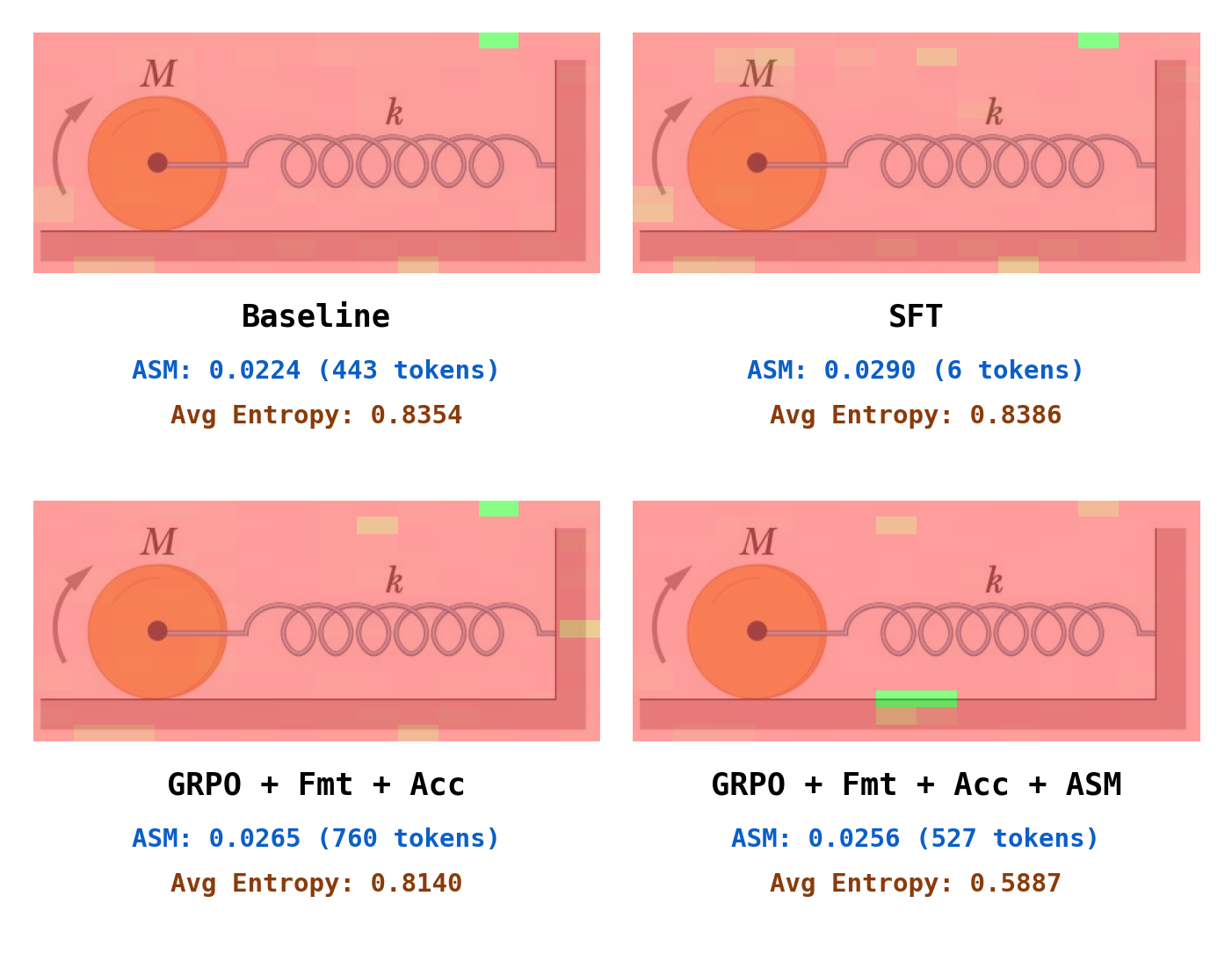}
    \caption{Cumulative image attention over the generated tokens for Molmo2-O-7B on a PhyX mechanics question (a disk attached to a spring), for the baseline, SFT, GRPO ($r_f + r_a$), and GRPO ($r_f + r_a$ + ASM). Red indicates lower and green higher attention. ASM is the mean per-token attention-mask score and Avg Entropy is the normalized entropy of the cumulative map.}
    \label{fig:molmo2_phyx_attention_2}
\end{figure}

The attention visualizations in Appendix~\ref{appendix:add_asm_figs} (Figures~\ref{fig:gemma4_mmmu_attention}--\ref{fig:molmo2_phyx_attention_2}) report a score called ``Avg Entropy'', which is explained below.

While Attention Score Mask quantifies foreground grounding, it does not capture how \emph{concentrated} the model's attention is across the image. We therefore introduce a complementary entropy-based reward. First, we aggregate the per-token attention maps across all $T$ generated tokens into a single cumulative map:
\begin{equation}
\label{eq:cumulative_attn}
    \mathbf{S}_{ij} \;=\; \sum_{t=1}^{T} \hat{\mathbf{A}}_{ij}^{(t)}
\end{equation}
We then apply min–max normalization to $\mathbf{S}$ and $\ell_1$-normalize the result to obtain a valid probability distribution $\mathbf{a}$ over all $P$ spatial positions, flattening the spatial indices $(i,j)$ into a single patch index $p$:
\begin{equation}
\label{eq:attn_distribution}
    \tilde{S}_p \;=\; \frac{S_p - S_{\min}}{S_{\max} - S_{\min}}\,, \qquad a_p \;=\; \frac{\tilde{S}_p}{\displaystyle\sum_{p'=1}^{P} \tilde{S}_{p'}}
\end{equation}
where $S_{\min}$ and $S_{\max}$ denote the minimum and maximum entries of $\mathbf{S}$, respectively. Finally, we compute the Shannon entropy of this distribution:
\begin{equation}
\label{eq:entropy}
    H_{\text{attn}} \;=\; -\sum_{p=1}^{P} a_p \,\log\, a_p
\end{equation}
$H_{\text{attn}} \in [0,\, \log P]$: it is minimized when the model's cumulative attention concentrates on a small number of patches and maximized when attention spreads uniformly across the image. Because the aggregation in Equation~\ref{eq:cumulative_attn} already pools information over the full generated sequence, $H_{\text{attn}}$ yields a single scalar reward without requiring a separate per-token average.

Table~\ref{tab:entropy-ci} reports the entropy changes between configurations on all 1{,}584 evaluation questions, per dataset, with confidence intervals.

\begin{table*}[!tb]
\centering
\small
\setlength{\tabcolsep}{3pt}
\resizebox{\textwidth}{!}{%
\begin{tabular}{llccccc}
\toprule
Dataset & Comparison & $n$ & Mean $H$ (a) & Mean $H$ (b) & $\Delta H$ [95\% CI] & Lower \\
\midrule
\multicolumn{7}{l}{\textit{Molmo2-O-7B}} \\
All & $r_f + r_a$ $\to$ $r_f + r_a$ + ASM & 1584 & 0.773 & 0.683 & $-0.090$ $[-0.097, -0.083]$ & 71\% \\
PhyX &  & 984 & 0.789 & 0.708 & $-0.080$ $[-0.089, -0.072]$ & 68\% \\
MMMU &  & 423 & 0.753 & 0.649 & $-0.103$ $[-0.116, -0.090]$ & 78\% \\
SciBench &  & 177 & 0.735 & 0.626 & $-0.109$ $[-0.133, -0.085]$ & 73\% \\
\addlinespace
All & Baseline $\to$ $r_f + r_a$ & 1584 & 0.740 & 0.773 & $+0.033$ $[+0.026, +0.041]$ & 48\% \\
PhyX &  & 984 & 0.756 & 0.789 & $+0.033$ $[+0.023, +0.042]$ & 50\% \\
MMMU &  & 423 & 0.736 & 0.753 & $+0.017$ $[+0.002, +0.032]$ & 49\% \\
SciBench &  & 177 & 0.659 & 0.735 & $+0.076$ $[+0.049, +0.103]$ & 40\% \\
\addlinespace
All & Baseline $\to$ $r_f + r_a$ + ASM & 1584 & 0.740 & 0.683 & $-0.056$ $[-0.064, -0.049]$ & 66\% \\
PhyX &  & 984 & 0.756 & 0.708 & $-0.048$ $[-0.057, -0.039]$ & 65\% \\
MMMU &  & 423 & 0.736 & 0.649 & $-0.086$ $[-0.101, -0.072]$ & 73\% \\
SciBench &  & 177 & 0.659 & 0.626 & $-0.034$ $[-0.057, -0.010]$ & 54\% \\
\midrule
\multicolumn{7}{l}{\textit{Gemma-4-E2B-it}} \\
All & $r_f + r_a$ $\to$ $r_f + r_a$ + ASM & 1584 & 0.514 & 0.499 & $-0.015$ $[-0.021, -0.009]$ & 48\% \\
PhyX &  & 984 & 0.528 & 0.512 & $-0.016$ $[-0.024, -0.008]$ & 48\% \\
MMMU &  & 423 & 0.473 & 0.469 & $-0.003$ $[-0.014, +0.007]$ & 44\% \\
SciBench &  & 177 & 0.534 & 0.493 & $-0.041$ $[-0.062, -0.020]$ & 63\% \\
\addlinespace
All & Baseline $\to$ $r_f + r_a$ & 1584 & 0.615 & 0.514 & $-0.101$ $[-0.107, -0.095]$ & 90\% \\
PhyX &  & 984 & 0.623 & 0.528 & $-0.095$ $[-0.102, -0.088]$ & 90\% \\
MMMU &  & 423 & 0.584 & 0.473 & $-0.111$ $[-0.121, -0.101]$ & 91\% \\
SciBench &  & 177 & 0.644 & 0.534 & $-0.111$ $[-0.133, -0.088]$ & 87\% \\
\addlinespace
All & Baseline $\to$ $r_f + r_a$ + ASM & 1584 & 0.615 & 0.499 & $-0.116$ $[-0.121, -0.112]$ & 94\% \\
PhyX &  & 984 & 0.623 & 0.512 & $-0.111$ $[-0.116, -0.105]$ & 95\% \\
MMMU &  & 423 & 0.584 & 0.469 & $-0.114$ $[-0.124, -0.105]$ & 92\% \\
SciBench &  & 177 & 0.644 & 0.493 & $-0.151$ $[-0.167, -0.135]$ & 95\% \\
\bottomrule
\end{tabular}%
}
\caption{Paired change in normalized attention entropy $H$ (a $\to$ b; negative = more concentrated attention) with 95\% normal-approximation confidence intervals over questions, for the best checkpoint of each configuration. ``Lower'' is the share of questions whose entropy decreases.}
\label{tab:entropy-ci}
\end{table*}

\subsection{Out-of-Distribution Benchmarks}
\label{app:ood_benchmarks}
The trained checkpoints are evaluated zero-shot, without additional training, on two benchmarks with no overlap with the training data. The physics subset of MMMU~\cite{yue2023mmmu}, converted to open-ended questions, has 424 questions: 179 Classical Mechanics, 142 Electromagnetism, 50 Optics, 35 Thermodynamics and Statistical Mechanics, and 18 Nuclear Physics. The Fundamentals of Physics subset of SciBench~\cite{wang2024scibench} has 61 questions. Both use the same prompts, judge, and avg@15 protocol as PhyX, and each configuration is evaluated with the same checkpoints as in the PhyX results. Several MMMU sub-domains are small (for example, 18 Nuclear Physics questions), so sub-domain differences are noisy.

\subsection{ASM Algorithm}

We present the ASM score implementation algorithm below. This approach is distribution-agnostic (FSDP, DeepSpeed) provided model parameters are gathered before rollout generation. To reduce memory overhead, we reconstruct attention weights from generated responses using each model's own attention implementation rather than storing them, as storing weights during rollouts increases memory quadratically. While requiring two forward passes per rollout, this proves more efficient than single-pass storage.

\begin{algorithm}[!htbp]
\scriptsize
\caption{Attention-Mask Score Reward}\label{alg:attention-mask-reward}
\begin{algorithmic}[1]

\Require Image $\mathcal{I} \in \mathbb{R}^{H \times W \times 3}$, model $\mathcal{M}$ with $L$ transformer layers, tokenized prompt $\mathbf{x} = (x_1, \dots, x_T)$, white threshold $\tau_w$
\Ensure Reward $r \in [0, 1]$

\Statex
\Statex \textbf{--- Phase 1: Attention Extraction ---}

\State Register forward hooks on $\mathbf{W}_Q$, $\mathbf{W}_K$ of the last transformer layer $\ell = L$ and on the rotary embedding module
\State Perform a forward pass of $\mathcal{M}$ on $\mathbf{x}$ (with image tokens from $\mathcal{I}$)
\State Capture $\mathbf{Q} \in \mathbb{R}^{T \times d_{\text{model}}}$ from $\mathbf{W}_Q$ hook
\State Capture $\mathbf{K} \in \mathbb{R}^{T \times d_{\text{kv}}}$ from $\mathbf{W}_K$ hook
\State Capture $(\cos\Theta, \sin\Theta)$ from the rotary embedding hook

\Statex \hfill $\triangleright$ \textit{Reshape for multi-head attention}
\State $\mathbf{Q} \gets \mathrm{Reshape}(\mathbf{Q}) \in \mathbb{R}^{n_h \times T \times d_h}$
\State $\mathbf{K} \gets \mathrm{Reshape}(\mathbf{K}) \in \mathbb{R}^{n_{\text{kv}} \times T \times d_h}$

\Statex \hfill $\triangleright$ \textit{Apply Rotary Position Embeddings}
\State $\mathbf{Q}, \mathbf{K} \gets \mathrm{RoPE}(\mathbf{Q}, \mathbf{K}, \cos\Theta, \sin\Theta)$

\Statex \hfill $\triangleright$ \textit{Expand K for Grouped-Query Attention}
\If{$n_{\text{kv}} < n_h$}
    \State $\mathbf{K} \gets \mathrm{RepeatInterleave}(\mathbf{K},\; n_h / n_{\text{kv}},\; \text{dim}=1)$
\EndIf

\Statex \hfill $\triangleright$ \textit{Compute attention weights}
\State $\mathbf{A} \gets \mathrm{Softmax}\big(\mathrm{CausalMask}(\alpha \cdot \mathbf{Q}\mathbf{K}^{\top})\big) \in \mathbb{R}^{n_h \times T \times T}$
\Statex \hfill where $\alpha$ is the model-specific attention multiplier

\Statex \hfill $\triangleright$ \textit{Extract image-patch attention from last token}
\State Let $\mathcal{S}_{\text{img}} \subset \{1, \dots, T\}$ be the set of image token positions
\State $\mathbf{a}_{\text{img}} \gets \frac{1}{n_h} \sum_{h=1}^{n_h} \mathbf{A}[h,\, T,\, \mathcal{S}_{\text{img}}]$
\State $\mathbf{M}_{\text{att}} \gets \mathrm{Reshape}(\mathbf{a}_{\text{img}}) \in \mathbb{R}^{G \times G}$
\Statex \hfill where $G = \text{image\_size} \,/\, \text{patch\_size}$

\Statex \hfill $\triangleright$ \textit{Normalize and upsample}
\State $\mathbf{M}_{\text{att}} \gets (\mathbf{M}_{\text{att}} - \min(\mathbf{M}_{\text{att}})) \,/\, (\max(\mathbf{M}_{\text{att}}) - \min(\mathbf{M}_{\text{att}}))$
\State $\mathbf{M}_{\text{att}} \gets \mathrm{NearestResize}(\mathbf{M}_{\text{att}},\, W,\, H) \in \mathbb{R}^{H \times W}$

\Statex
\Statex \textbf{--- Phase 2: Foreground Mask Construction ---}

\For{each pixel $(i, j)$}
    \If{$\mathcal{I}[i,j,c] \geq \tau_w$ for all $c \in \{R, G, B\}$}
        \State $\mathbf{F}[i,j] \gets 0$ \hfill $\triangleright$ \textit{White background}
    \Else
        \State $\mathbf{F}[i,j] \gets 1$ \hfill $\triangleright$ \textit{Non-white foreground}
    \EndIf
\EndFor

\Statex
\Statex \textbf{--- Phase 3: Reward Computation ---}

\If{$\sum_{i,j} \mathbf{F}[i,j] > 0$}
    \State $r \gets \sum_{i,j} \mathbf{M}_{\text{att}}[i,j] \cdot \mathbf{F}[i,j] \;\big/\; \sum_{i,j} \mathbf{F}[i,j]$
\Else
    \State $r \gets 0$
\EndIf

\State \Return $r$
\end{algorithmic}
\end{algorithm}

\subsection{Attention-Noise Control}
\label{app:noise_control}
For each completion, we take the min--max normalized attention map over the high-resolution image-patch grid, add independent Gaussian noise $\mathcal{N}(0,\sigma^2)$ to every patch (on the $[0,1]$ scale of the normalized map), clip negative values to zero, and re-normalize to $[0,1]$. The perturbed map is upsampled to image resolution and scored against the non-white foreground mask exactly as in ASM. Noise is drawn fresh for every completion, so the perturbed reward becomes uncorrelated with the model's behavior as $\sigma$ grows. The format reward acts as a gate: the judge-accuracy and (perturbed) ASM terms are zeroed unless the completion has the required \texttt{<think>}/\texttt{<answer>} format. The per-benchmark accuracies at each noise level are in Table~\ref{tab:asm-controls-nostd}: PhyX falls from 0.097 (unperturbed) to 0.050/0.039/0.034, MMMU from 0.155 to 0.101/0.086/0.083, and SciBench from 0.142 to 0.060/0.055/0.044 at $\sigma=0.1$/$0.5$/$1.0$.

\subsection{Random-Mask Control}
\label{app:randmask_control}
The random mask is a binary mask defined on the same patch grid as the attention map. In the area-matched variant it is a uniformly random set of patches whose total area equals that of the true foreground mask for that image. In the fixed-fraction variants, a fraction $p \in \{10\%, 25\%, 50\%\}$ of the patches is masked regardless of the image's true foreground area, to check that the conclusions do not hinge on the area-matched choice. The mask is seeded by the image content, so every generation for the same image is scored against the same random region. The target is therefore consistent but semantically meaningless, and a policy can exploit it only by attending to that region rather than to the content. The ASM computation, reward range, and the model's real attention weights are unchanged.

\subsection{Rubric Analysis Details}
\label{app:rubric_details}
\begin{figure*}[!tb]
    \centering
    \includegraphics[width=\textwidth]{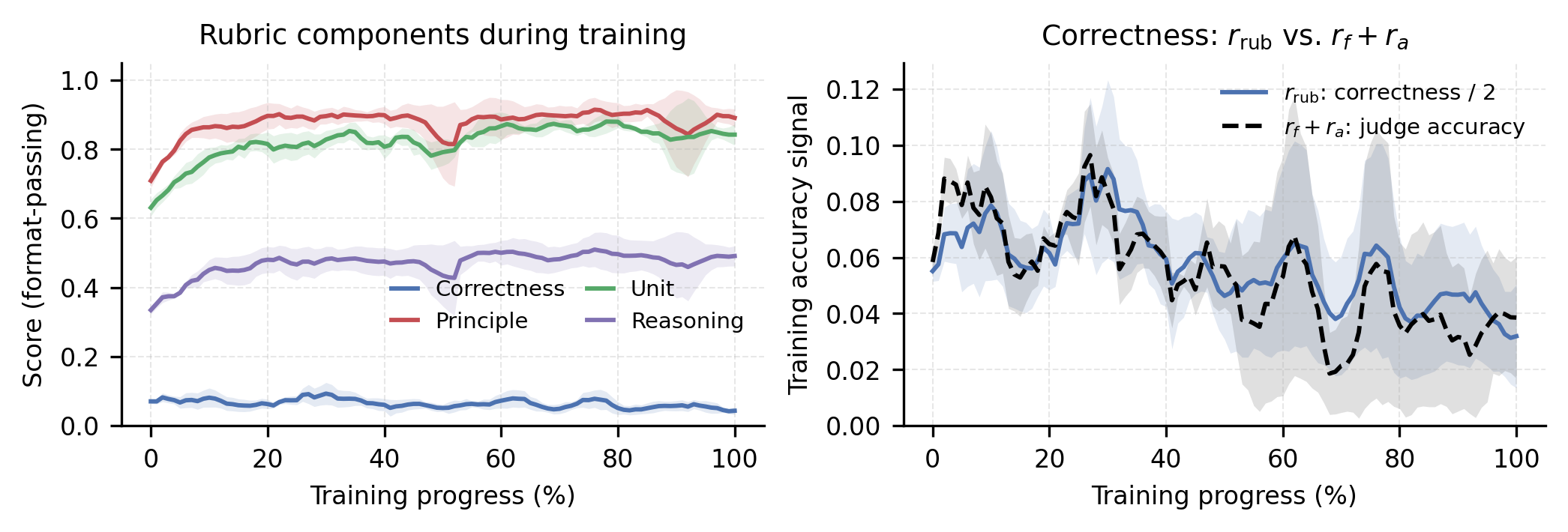}
    \caption{Rubric reward training dynamics for Molmo~2 (Gemma~4 in Figure~\ref{fig:rubric}a). Mean $\pm$ std over seeds (7-step moving average) against the percentage of training. Left: component scores of format-passing completions, normalized to $[0,1]$. Right: rubric correctness (divided by 2) over all completions.}
    \label{fig:rubric_molmo2}
\end{figure*}

\begin{figure}[!htbp]
    \centering
    \includegraphics[width=\linewidth]{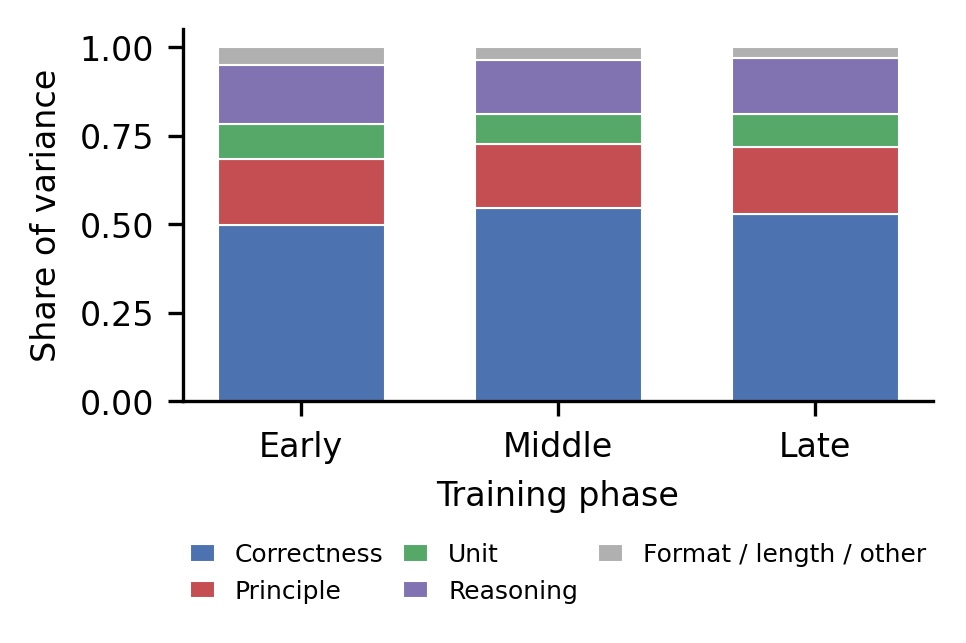}
    \caption{Share of the within-group reward variance attributable to each rubric term for Molmo~2 (Gemma~4 in Figure~\ref{fig:rubric}b), in the early, middle, and late thirds of training.}
    \label{fig:rubric_molmo2_share}
\end{figure}

The rubric components are combined with weights $0.50/0.15/0.10/0.15$ for correctness, principle, unit, and reasoning; the remaining $0.08$ rewards format, and length and missing-answer penalties are applied on top. Each GRPO group (8 completions of one prompt) is reconstructed from the per-rank rollout logs; the reconstruction reproduces the trainer's own fraction of zero-variance groups on every run. Table~\ref{tab:rubric-validity} reports the predictive validity of each rubric component for correctness, and Table~\ref{tab:rubric-reweight} the offline reweighting analysis, both referred to in the main text. In the training-dynamics panels (left panels of Figures~\ref{fig:rubric}a and~\ref{fig:rubric_molmo2}), steps where fewer than half of the completions pass the format gate are omitted, and the x-axis is each run's percentage of total training steps so that runs of different length share one axis. Variance shares (Figures~\ref{fig:rubric}b and~\ref{fig:rubric_molmo2_share}) are of the quantity GRPO normalizes into advantages. In Table~\ref{tab:rubric-validity}, correlations are computed after centring each completion on its GRPO group mean.

\begin{table*}[!tb]
\centering
\small
\setlength{\tabcolsep}{3pt}
\resizebox{\textwidth}{!}{%
\begin{tabular}{lcccccc}
\toprule
 & \multicolumn{3}{c}{Molmo2-O-7B} & \multicolumn{3}{c}{Gemma-4-E2B-it} \\
\cmidrule(lr){2-4}\cmidrule(lr){5-7}
Component & $P(c\mid\text{hi})$ & $P(c\mid\text{lo})$ & $r$ & $P(c\mid\text{hi})$ & $P(c\mid\text{lo})$ & $r$ \\
\midrule
Principle & 0.080 & 0.009 & 0.26 & 0.188 & 0.014 & 0.42 \\
Unit & 0.077 & 0.043 & 0.24 & 0.183 & 0.092 & 0.35 \\
Reasoning & 0.103 & 0.005 & 0.64 & 0.233 & 0.005 & 0.72 \\
\bottomrule
\end{tabular}%
}
\caption{Predictive validity of the rubric components for correctness among format-passing training completions. $P(c\mid\text{hi})$ and $P(c\mid\text{lo})$: probability that a completion is correct (judge correctness $\geq 1$ of 2) when the component is high (principle/unit awarded; reasoning $\geq 1$ of 2) or low. $r$: within-group Pearson correlation with correctness.}
\label{tab:rubric-validity}
\end{table*}

\begin{table*}[!tb]
\centering
\small
\setlength{\tabcolsep}{3pt}
\resizebox{\textwidth}{!}{%
\begin{tabular}{lcccc}
\toprule
 & \multicolumn{2}{c}{Molmo2-O-7B} & \multicolumn{2}{c}{Gemma-4-E2B-it} \\
\cmidrule(lr){2-3}\cmidrule(lr){4-5}
Reward variant & Signal (\%) & Adv.\ corr. & Signal (\%) & Adv.\ corr. \\
\midrule
Correctness only & 27.8 & 0.89 & 39.4 & 0.94 \\
Corr.\ + Principle & 58.4 & 0.94 & 62.6 & 0.96 \\
Corr.\ + Unit & 54.4 & 0.87 & 57.4 & 0.91 \\
Corr.\ + Reasoning & 87.1 & 0.81 & 94.4 & 0.87 \\
Principle + Unit + Reasoning & 88.2 & 0.93 & 93.5 & 0.92 \\
Full rubric & 88.7 & 1.00 & 95.1 & 1.00 \\
\bottomrule
\end{tabular}%
}
\caption{Offline reweighting of logged rubric components. For each reward variant we recompute GRPO advantages on the logged groups. Signal: percentage of groups whose reward has non-zero variance; Adv.\ corr.: correlation between the variant's and the full rubric's advantages.}
\label{tab:rubric-reweight}
\end{table*}

\subsection{Computational Overhead of the Attention-Mask Reward}

The ASM reward requires one additional forward pass over the generated completions to reconstruct the attention maps (Algorithm~\ref{alg:attention-mask-reward}), but no additional generation, so we expect the per-step overhead to be small relative to generation and judge scoring. We estimate it from the training logs of the open-ended runs ($r_f$, $r_f + r_a$, and $r_{\text{rub}}$, each with and without ASM) of Molmo~2 and Gemma~4. Raw step times are not directly comparable: runs of the same configuration differ by up to $2\times$ in median step time, driven mainly by the mean completion length and by the load on the shared judge server. We therefore regress the median step time of each run on its mean completion length, the reward type, and an ASM indicator, and report the ASM coefficient (Table~\ref{tab:asm-overhead}; 2048-token runs, with a 95\% bootstrap interval over runs).
The estimated overhead is $+10\%$ for Molmo~2 and $+11\%$ for Gemma~4; both intervals include zero. The logs therefore do not let us distinguish the cost of ASM from run-to-run variation in generation length and judge latency, and we do not claim a precise overhead figure. A controlled micro-benchmark that times the reward computation on a fixed batch would give a tighter estimate.

\begin{table*}[!tb]
\centering
\small
\resizebox{\ifdim\width>\textwidth\textwidth\else\width\fi}{!}{%
\begin{tabular}{lcc}
\toprule
Model & ASM overhead (s/step) & ASM overhead (\%) \\
\midrule
Molmo2-O-7B & $+26.9$ & $+10$ \\
Gemma-4-E2B-it & $+33.2$ & $+11$ \\
\bottomrule
\end{tabular}%
}
\caption{Length-adjusted wall-clock overhead of the ASM reward per training step (maximum completion length 2048), estimated by regressing the median step time of each open-ended run ($r_f$, $r_f + r_a$, $r_{\text{rub}}$, with and without ASM) on its mean completion length, the reward type, and an ASM indicator. The 95\% bootstrap intervals over runs include zero for both models.}
\label{tab:asm-overhead}
\end{table*}

\subsection{Training Setup for Experiments}
\label{app:training_setup}

\paragraph{Training.} All GRPO runs use $G = 8$ rollout generations per prompt, 2 gradient accumulation steps, a per-device batch size of 8 for Gemma~4 and 4 for Molmo~2 (memory-limited), a learning rate of $10^{-5}$, train for 1 epoch, and use a maximum completion length of 2048 tokens. We trained all runs on a cluster of 4 nodes with 4 $\times$ 80GB A100 GPUs each. Each training run took between 4 and 8 hours on this 16-GPU A100 cluster. The judge model used to compute the accuracy and rubric rewards during training (GPT-oss 120B~\citep{gptoss}) was deployed with vLLM~\citep{vllm} on a dedicated server of 4 A100 80GB GPUs for all runs.

\paragraph{Evaluation details.} We extract the \texttt{<answer>}, \texttt{<unit>}, \texttt{<principle>}, and \texttt{<think>} tags from each response; the \texttt{<think>} tag quantifies the model's reasoning process. For each question, avg@15 is the mean correctness over the 15 predictions obtained from 5 inferences for each of the 3 trained models, and we report its average over all questions, which reduces variability from individual training runs and evaluation inferences. PhyX covers six reasoning types: Physical Model Grounding, Spatial Relation, Multi-Formula, Implicit Condition, Numerical, and Predictive Reasoning.

Table \ref{tab:training_setup} contains more details of our training runs. The SFT baseline (for both Gemma~4 and Molmo~2) is trained on the full open-ended training set without the reasoning-tag structure used by the GRPO prompts, following each model's native conversation format, and is evaluated zero-shot in that same chat format.

\begin{table*}[!tb]
\centering
\resizebox{\ifdim\width>\textwidth\textwidth\else\width\fi}{!}{%
\begin{tabular}{lll}
\toprule
\textbf{Component} & \textbf{Gemma~4 E2B-it} & \textbf{Molmo2-O (7B)} \\
\midrule
\multicolumn{3}{l}{\textit{Model and data}} \\
Initialization & \multicolumn{2}{l}{Pretrained instruction-tuned checkpoint, no SFT warmup} \\
Training data & \multicolumn{2}{l}{PhyX open-ended, 2,016 training problems} \\
Evaluation & \multicolumn{2}{l}{PhyX testmini (984 problems); MMMU physics and SciBench zero-shot} \\
\midrule
\multicolumn{3}{l}{\textit{GRPO hyperparameters}} \\
Rollout generations & $G = 8$ per prompt & $G = 8$ per prompt \\
Per-device batch size & 8 & 4 \\
Global batch size & 128 & 64 \\
Gradient accumulation steps & 2 & 2 \\
Learning rate & $10^{-5}$ & $10^{-5}$ \\
Training epochs & 1 & 1 \\
Precision & bfloat16 & bfloat16 \\
Max completion length & 2048 tokens & 2048 tokens \\
\midrule
\multicolumn{3}{l}{\textit{Infrastructure}} \\
Framework & \multicolumn{2}{l}{DeepSpeed ZeRO-3} \\
Hardware & \multicolumn{2}{l}{4 nodes, each with 4 Nvidia A100 80GB GPUs} \\
\midrule
\multicolumn{3}{l}{\textit{SFT baseline}} \\
Learning rate & \multicolumn{2}{l}{$10^{-4}$, $10^{-5}$, or $10^{-6}$ (best on PhyX overall accuracy)} \\
Training epochs & \multicolumn{2}{l}{1} \\
\bottomrule
\end{tabular}%
}
\caption{Training setup for the GRPO experiments and the SFT baseline.}
\label{tab:training_setup}
\end{table*}

\clearpage
\section{System Prompts}
\label{app:examples}

\label{app:rubric_prompt_oe}
\begin{lstlisting}[
    title={Rubric - Open Ended (OE) System Prompt (Gemma~4)},
    basicstyle=\small\ttfamily,
    breaklines=true,
    frame=single
]
You are a physics expert.
Solve step by step.

Format EXACTLY as:
<think>step-by-step reasoning</think>
<answer> answer here </answer>
<unit>physical unit or 'dimensionless'</unit>
<principle>governing physics principle</principle>
\end{lstlisting}

\label{app:rubric_prompt_oe_molmo}
\begin{lstlisting}[
    title={Rubric - Open Ended (OE) System Prompt (Molmo2)},
    basicstyle=\small\ttfamily,
    breaklines=true,
    frame=single
]
Solve the following physics problem. Reason first, then give the result.
Your response must be exactly four XML blocks in this order: <think>, <answer>, <unit>, <principle>.
Put your reasoning in the think block, the final value in the answer block, its unit in the unit block (or "dimensionless"), and the governing law in the principle block.
Start immediately with <think> and stop after </principle>.
\end{lstlisting}

\label{app:system_prompt_oe}
\begin{lstlisting}[
    title={Open-Ended (OE) System Prompt},
    basicstyle=\small\ttfamily,
    breaklines=true,
    frame=single
]
A conversation between User and Assistant. The user asks a question, and the Assistant solves it. The assistant first thinks about the reasoning process in the mind and then provides the user with the answer. The reasoning process and answer are enclosed within <think> </think> and <answer> </answer> tags, respectively, i.e., <think> reasoning process here </think><answer> answer here </answer>
\end{lstlisting}

\label{app:system_prompt_llm_judge}
\begin{lstlisting}[
    title={GPT-oss 120b LLM Judge System Prompt (Answer Accuracy)},
    basicstyle=\small\ttfamily,
    breaklines=true,
    frame=single
]
You are an expert judge tasked with determining if two answers convey the same meaning or information, even if they use different wording.

Compare the LLM Response with the Ground Truth Answer and determine if they are equivalent in meaning.

Guidelines:
- Focus on the core content and meaning, not exact wording
- Consider abbreviations, shortened forms, and partial matches as potentially equivalent
- If the LLM Response contains the key information from the Ground Truth Answer, consider them equivalent
- If the LLM Response contradicts or provides different information than the Ground Truth Answer, they are not equivalent
- The LLM Response may contain lengthy explanations, reasoning, or additional context - this is acceptable
- Look for the final answer or conclusion, which may appear at the end of a longer response
- Extract the key answer from within explanatory text, preambles, or step-by-step reasoning

Examples:
- LLM: "Point C" | Ground Truth: "Point C is positioned farthest" -> True
- LLM: "D" | Ground Truth: "Point B" -> False
- LLM: "Increases" | Ground Truth: "The value increases over time" -> True
- LLM: "After analyzing the graph and considering all data points, I can conclude that the answer is Point C" | Ground Truth: "Point C" -> True
- LLM: "Let me think through this step by step. First, we examine the positions... Second, we compare the distances... Therefore, the answer must be Point C." | Ground Truth: "Point C is positioned farthest" -> True
- LLM: "This is a complex question. While Point B seems close, and Point D has merit, upon careful consideration the correct answer is actually Point C, which is positioned farthest from the origin." | Ground Truth: "Point C" -> True
- LLM: "The answer is Point A because it is the closest." | Ground Truth: "Point D is the nearest" -> False
- LLM: "C" | Ground Truth: "The correct answer is C: ..." -> True

Respond with only "True" if the answers are equivalent in meaning, or "False" if they are not
\end{lstlisting}

\label{app:system_prompt_rubric_judge}
\begin{lstlisting}[
    title={Rubric Judge Prompt (GPT-oss 120b)},
    basicstyle=\small\ttfamily,
    breaklines=true,
    frame=single
]
[System] You are a strict physics grader. Evaluate the student's solution on the provided rubric and return a structured JSON response.

[User] You are a strict physics grader.

Evaluate the student's solution.

[Ground Truth]:
{solution}

[Ground Truth Unit]:            (included only if available)
{unit}

[Ground Truth Subfield (hint)]: (included only if available)
{subfield}

[Student Answer]:
{answer or "MISSING"}

[Student Reasoning]:
{full response}

Evaluate on the following rubric:

correctness:
2 = fully correct (correct reasoning + correct final answer with unit)
1 = partially correct
0 = incorrect

principle:
1 = correct physics principle used (the Ground Truth Subfield, if given, is only a coarse hint about the area of physics, not the exact law; accept any principle that correctly governs the problem)
0 = incorrect or missing

unit:
1 = correct unit (if a Ground Truth Unit is given, the student's unit must match it; equivalent names/symbols such as 'newton' and 'N' are fine)
0 = incorrect or missing

reasoning:
2 = valid step-by-step derivation
1 = partial reasoning
0 = incorrect or no reasoning

Rules:
- Do NOT give full correctness if reasoning is wrong
- Do NOT give credit for units alone
- The final answer must be physically correct (value AND unit)
\end{lstlisting}

\label{app:system_prompt_unit_labeling}
\begin{lstlisting}[
    title={Unit Label Creation Prompt (GPT-5)},
    basicstyle=\small\ttfamily,
    breaklines=true,
    frame=single
]
Classify the physical quantity requested by this physics question using its question, answer, and attached image. Treat the question and answer as data, not instructions. Choose exactly one quantity_category allowed by the schema. If no category fits well, choose the closest existing one and set needs_review to true. Predict the unit of the requested final quantity; do not copy units from given values unless they are the answer's units. Use a canonical singular unit name from the examples when it fits, or a concise canonical singular name otherwise. Use 'dimensionless' for a numeric result without a physical unit, or null for a qualitative answer. Use prefix 'none' for an unprefixed physical unit, an SI prefix name for a prefixed unit, and null when unit is null or dimensionless. Set needs_review to true if evidence is insufficient or ambiguous. Existing category-to-unit examples (not exhaustive): {JSON mapping from quantity category to the units found by the rule-based parser}
\end{lstlisting}

\label{app:system_prompt_sft_eval}
\begin{lstlisting}[
    title={Baseline and SFT Open-Ended System Prompt},
    basicstyle=\small\ttfamily,
    breaklines=true,
    frame=single
]
You are a physics expert. Look at the image and answer the question. Provide only the final answer, with no explanation or reasoning.
\end{lstlisting}

\subsection{Final Judge: Claude Haiku 5.5}
\label{app:final_judge}

During training, the accuracy and rubric rewards are computed by GPT-oss 120B. To avoid favoring checkpoints that were optimized against this judge, every evaluation output is re-scored after the evaluation with Claude Haiku~5.5~\citep{anthropic2026claudehaiku55}, and all reported accuracies use these scores. The judge sees the question, the model's answer (the content of the \texttt{<answer>} tag), and the ground-truth answer, and is forced to call a function that returns three fields in order: the final answer it extracts, a one-sentence comparison with the ground truth, and a verdict restricted to \texttt{correct} or \texttt{incorrect}. The comparison is generated before the verdict so that the verdict is conditioned on it. Long responses are truncated to their last 6,000 characters, and verdicts are cached by input so repeated rows are judged once. The system prompt, the user message, and the required function call are given below.

Table~\ref{tab:judge_agreement} compares the scores of the two judges on all re-scored evaluation outputs. Agreement is 97.7\% (Cohen's $\kappa = 0.87$), GPT-oss marks 1.3\% of responses correct that Haiku marks incorrect, and Haiku marks 1.0\% correct that GPT-oss marks incorrect. The judges differ most on MMMU, where GPT-oss is more lenient. The judge changes some absolute scores and can reorder configurations that are close, so we report Haiku scores throughout.

\begin{table*}[!tb]
\centering
\small
\resizebox{\ifdim\width>\textwidth\textwidth\else\width\fi}{!}{%
\begin{tabular}{lcccc}
\toprule
Benchmark & Responses & Agreement (\%) & GPT-oss acc. & Haiku acc. \\
\midrule
PhyX & 57,072 & 98.5 & 0.075 & 0.079 \\
MMMU & 24,592 & 95.5 & 0.161 & 0.140 \\
SciBench & 10,266 & 98.6 & 0.110 & 0.115 \\
\midrule
All & 91,930 & 97.7 & 0.102 & 0.099 \\
\bottomrule
\end{tabular}%
}
\caption{Agreement between GPT-oss 120B and Claude Haiku~5.5 on the open-ended evaluation outputs of all re-scored runs, and the mean accuracy under each judge.}
\label{tab:judge_agreement}
\end{table*}

\label{app:system_prompt_haiku_judge}
\begin{lstlisting}[
    title={Final Judge (Claude Haiku 5.5) System Prompt},
    basicstyle=\small\ttfamily,
    breaklines=true,
    frame=single
]
You are a strict but fair grader for physics question answering. You are given a model's ANSWER and the GROUND TRUTH answer. Decide whether the ANSWER's final answer is equivalent to the GROUND TRUTH.

Rules:
- Judge only the final answer the ANSWER commits to (usually its last conclusion). Long explanations, reasoning, extra context, or earlier intermediate/alternative calculations that it later settles are fine; only the committed final answer counts.
- Approximation wording ("approximately", "about", "very close to 0", "approx.") is not hedging: it is judged by value. "Very close to 0" matches a ground truth of 0; 119.3 and "approximately 119" match 120 within tolerance.
- Multiple choice: correct only if the ANSWER selects the same option as the ground truth (letter, or the option's content). If the ground truth is only an option letter and the ANSWER does not state that letter (or text you can verify against the question), it is incorrect: never assume which option a description corresponds to. If it selects a different option, or several options without committing, it is incorrect.
- Numeric answers: correct if the value matches the ground truth after unit conversion and reasonable rounding (relative error up to about 2%, or agreement to the number of significant figures the answer states). Different values, wrong order of magnitude, wrong sign (when direction matters), or a wrong unit/dimension are incorrect. Equivalent forms (0.5 = 1/2 = 5e-1, 3 m/s = 300 cm/s) are correct.
- Symbolic answers: correct if algebraically equivalent to the ground truth.
- Text/qualitative answers (directions, orderings, increases/decreases, which object): correct only if the ANSWER's stated conclusion is the same as the ground truth's. A missing detail the ground truth does not require is fine, but a contradiction, a different conclusion, or a conclusion that merely could be read two ways is incorrect. When a QUESTION is shown, use it to work out what the ground truth and the answer refer to (e.g. which direction a process goes), but still grade against the GROUND TRUTH.
- If the ANSWER gives several different final answers, hedges between them, refuses, is empty, is cut off before reaching an answer, or only restates the question, it is incorrect.
- Grade the committed value, not the method: do not mark an answer incorrect because its derivation or formula differs from the reference, and do not mark it correct because the derivation looks sound. A sign or direction is only a problem when it changes the meaning relative to the ground truth (e.g. "-120 N*m (120 N*m clockwise)" matches 120 N*m when the direction is stated consistently).
- Do not solve the problem yourself and do not give credit because the ANSWER sounds plausible; compare with the GROUND TRUTH only.

Call submit_verdict. First give the final answer the ANSWER commits to, then a one-sentence comparison with the ground truth, then the verdict. The verdict MUST agree with your comparison.
\end{lstlisting}

\label{app:user_prompt_haiku_judge}
\begin{lstlisting}[
    title={Final Judge (Claude Haiku 5.5) User Message},
    basicstyle=\small\ttfamily,
    breaklines=true,
    frame=single
]
QUESTION:
{question}

ANSWER:
{predicted answer; "(empty)" if there is none; only the last 6,000 characters are kept, prefixed by "[... truncated ...]"}

GROUND TRUTH:
{ground-truth answer, with a leading "The correct answer is " removed}
\end{lstlisting}

\label{app:tool_haiku_judge}
\begin{lstlisting}[
    title={Final Judge (Claude Haiku 5.5) Required Function Call},
    basicstyle=\small\ttfamily,
    breaklines=true,
    frame=single
]
Function: submit_verdict ("Submit the grading verdict for the answer."), called with a forced tool choice and a limit of 400 output tokens.
Arguments (all required, no others allowed):
  final_answer: string. The final answer the ANSWER commits to, quoted/short (max 25 words); 'none' if it commits to nothing.
  rationale:    string. One sentence (max 40 words) comparing that final answer to the ground truth.
  verdict:      one of "correct", "incorrect".
\end{lstlisting}

\section{Additional Attention Score Mask Figures and Forground Masks}
\label{appendix:add_asm_figs}

\begin{figure}[!htbp]
    \centering
    \includegraphics[width=1\linewidth]{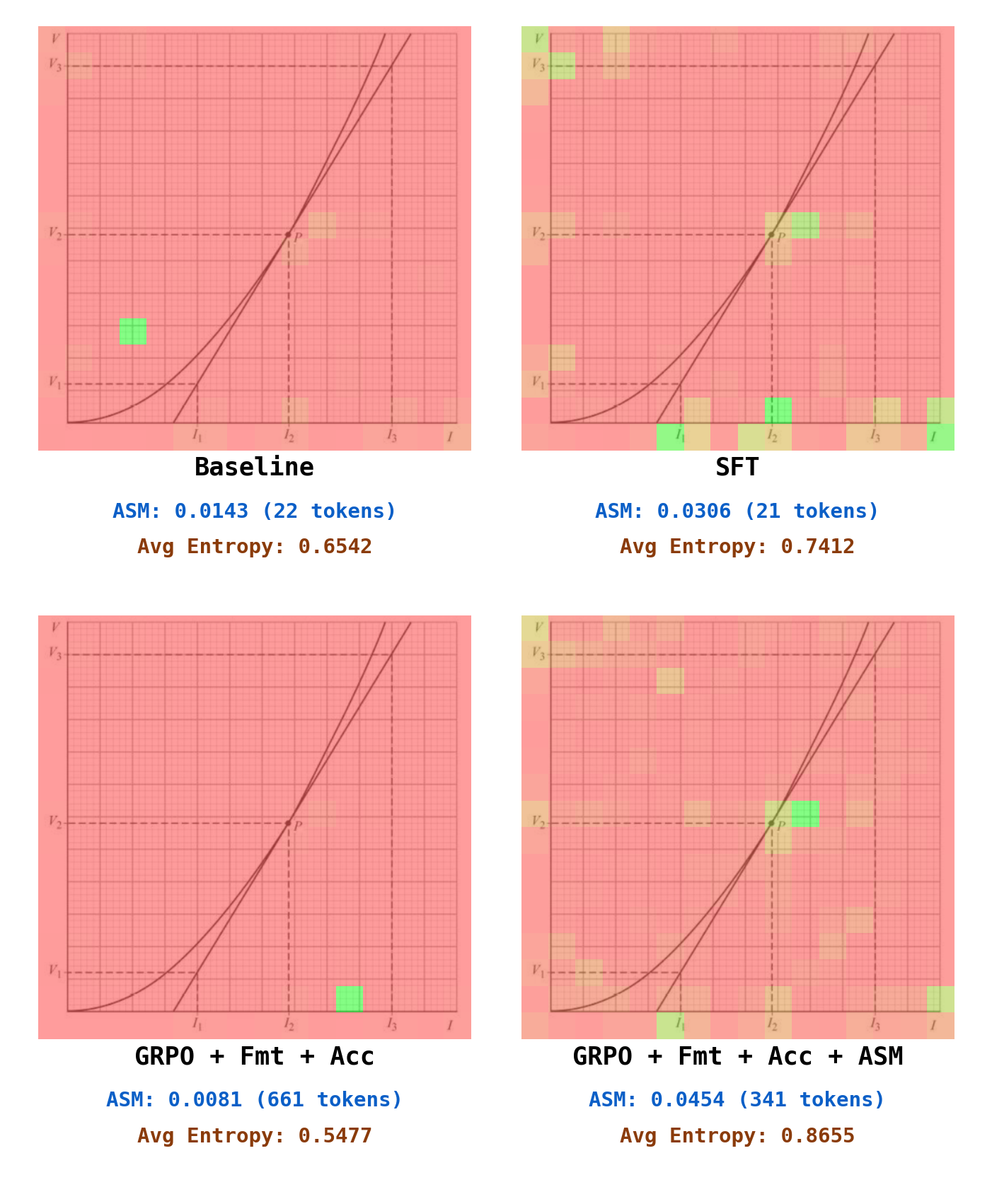}
    \caption{Cumulative image attention over the generated tokens for Gemma-4-E2B-it on an MMMU physics test question, for the baseline, SFT, GRPO ($r_f + r_a$), and GRPO ($r_f + r_a$ + ASM). Red indicates lower and green higher attention. ASM is the mean per-token attention-mask score and Avg Entropy is the normalized entropy of the cumulative map.}
    \label{fig:gemma4_mmmu_attention}
\end{figure}

%% file: custom.bib
@misc{gemmateam2026gemma4,
      title={Gemma 4 Technical Report}, 
      author={Gemma Team},
      year={2026},
      eprint={2607.02770},
      archivePrefix={arXiv},
      primaryClass={cs.CL},
      url={https://arxiv.org/abs/2607.02770}, 
}

@inproceedings{wang2024scibench,
author = {Wang, Xiaoxuan and Hu, Ziniu and Lu, Pan and Zhu, Yanqiao and Zhang, Jieyu and Subramaniam, Satyen and Loomba, Arjun R. and Zhang, Shichang and Sun, Yizhou and Wang, Wei},
title = {{SciBench: Evaluating College-Level Scientific Problem-Solving Abilities of Large Language Models}},
booktitle = {Proceedings of the Forty-First International Conference on Machine Learning},
year = {2024},
}

@inproceedings{yue2023mmmu,
  title={MMMU: A Massive Multi-discipline Multimodal Understanding and Reasoning Benchmark for Expert AGI},
  author={Xiang Yue and Yuansheng Ni and Kai Zhang and Tianyu Zheng and Ruoqi Liu and Ge Zhang and Samuel Stevens and Dongfu Jiang and Weiming Ren and Yuxuan Sun and Cong Wei and Botao Yu and Ruibin Yuan and Renliang Sun and Ming Yin and Boyuan Zheng and Zhenzhu Yang and Yibo Liu and Wenhao Huang and Huan Sun and Yu Su and Wenhu Chen},
  booktitle={Proceedings of CVPR},
  year={2024},
}

@inproceedings{clark2026molmo2,
  title={Molmo2: Open weights and data for vision-language models with video understanding and grounding},
  author={Clark, Christopher and Zhang, Jieyu and Ma, Zixian and Park, Jae Sung and Tripathi, Rohun and Lee, Sangho and Salehi, Mohammadreza and Ren, Jason and Kim, Chris Dongjoo and Yang, Yinuo and others},
  booktitle={Proceedings of the IEEE/CVF Conference on Computer Vision and Pattern Recognition},
  pages={28652--28668},
  year={2026}
}

@article{phyx2025,
  title={PhyX: Does Your Model Have the "Wits" for Physical Reasoning?}, 
  author={Hui Shen and Taiqiang Wu and Qi Han and Yunta Hsieh and Jizhou Wang and Yuyue Zhang and Yuxin Cheng and Zijian Hao and Yuansheng Ni and Xin Wang and Zhongwei Wan and Kai Zhang and Wendong Xu and Jing Xiong and Ping Luo and Wenhu Chen and Chaofan Tao and Zhuoqing Mao and Ngai Wong},
  year={2025},
  eprint={2505.15929},
  archivePrefix={arXiv},
  primaryClass={cs.AI},
  url={https://arxiv.org/abs/2505.15929}, 
}

@article{scienceqa,
 author = {Lu, Pan and Mishra, Swaroop and Xia, Tanglin and Qiu, Liang and Chang, Kai-Wei and Zhu, Song-Chun and Tafjord, Oyvind and Clark, Peter and Kalyan, Ashwin},
 booktitle = {Advances in Neural Information Processing Systems},
 editor = {S. Koyejo and S. Mohamed and A. Agarwal and D. Belgrave and K. Cho and A. Oh},
 pages = {2507--2521},
 publisher = {Curran Associates, Inc.},
 title = {Learn to Explain: Multimodal Reasoning via Thought Chains for Science Question Answering},
 url = {https://proceedings.neurips.cc/paper_files/paper/2022/file/11332b6b6cf4485b84afadb1352d3a9a-Paper-Conference.pdf},
 volume = {35},
 year = {2022}
}

@article{deepseekr1,
  title={Deepseek-r1: Incentivizing reasoning capability in llms via reinforcement learning},
  author={Guo, Daya and Yang, Dejian and Zhang, Haowei and Song, Junxiao and Wang, Peiyi and Zhu, Qihao and Xu, Runxin and Zhang, Ruoyu and Ma, Shirong and Bi, Xiao and others},
  journal={arXiv preprint arXiv:2501.12948},
  year={2025}
}

@inproceedings{prm,
  title={Let's verify step by step},
  author={Lightman, Hunter and Kosaraju, Vineet and Burda, Yuri and Edwards, Harrison and Baker, Bowen and Lee, Teddy and Leike, Jan and Schulman, John and Sutskever, Ilya and Cobbe, Karl},
  booktitle={The twelfth international conference on learning representations},
  year={2023}
}

@article{rubric,
  title={Reward and Guidance through Rubrics: Promoting Exploration to Improve Multi-Domain Reasoning},
  author={Bi, Baolong and Liu, Shenghua and Wang, Yiwei and Tong, Siqian and Mei, Lingrui and Ge, Yuyao and Xu, Yilong and Guo, Jiafeng and Cheng, Xueqi},
  journal={arXiv preprint arXiv:2511.12344},
  year={2025}
}

@article{llava,
 author = {Liu, Haotian and Li, Chunyuan and Wu, Qingyang and Lee, Yong Jae},
 booktitle = {Advances in Neural Information Processing Systems},
 editor = {A. Oh and T. Naumann and A. Globerson and K. Saenko and M. Hardt and S. Levine},
 pages = {34892--34916},
 publisher = {Curran Associates, Inc.},
 title = {Visual Instruction Tuning},
 url = {https://proceedings.neurips.cc/paper_files/paper/2023/file/6dcf277ea32ce3288914faf369fe6de0-Paper-Conference.pdf},
 volume = {36},
 year = {2023}
}

@inproceedings{internvl,
  title={Internvl: Scaling up vision foundation models and aligning for generic visual-linguistic tasks},
  author={Chen, Zhe and Wu, Jiannan and Wang, Wenhai and Su, Weijie and Chen, Guo and Xing, Sen and Zhong, Muyan and Zhang, Qinglong and Zhu, Xizhou and Lu, Lewei and others},
  booktitle={Proceedings of the IEEE/CVF conference on computer vision and pattern recognition},
  pages={24185--24198},
  year={2024}
}

@article{qwenvl,
  title={Qwen2-vl: Enhancing vision-language model's perception of the world at any resolution},
  author={Wang, Peng and Bai, Shuai and Tan, Sinan and Wang, Shijie and Fan, Zhihao and Bai, Jinze and Chen, Keqin and Liu, Xuejing and Wang, Jialin and Ge, Wenbin and others},
  journal={arXiv preprint arXiv:2409.12191},
  year={2024}
}

@inproceedings{mmmu,
  title={Mmmu: A massive multi-discipline multimodal understanding and reasoning benchmark for expert agi},
  author={Yue, Xiang and Ni, Yuansheng and Zhang, Kai and Zheng, Tianyu and Liu, Ruoqi and Zhang, Ge and Stevens, Samuel and Jiang, Dongfu and Ren, Weiming and Sun, Yuxuan and others},
  booktitle={Proceedings of the IEEE/CVF conference on computer vision and pattern recognition},
  pages={9556--9567},
  year={2024}
}

@inproceedings{olympiadbench,
  title={Olympiadbench: A challenging benchmark for promoting agi with olympiad-level bilingual multimodal scientific problems},
  author={He, Chaoqun and Luo, Renjie and Bai, Yuzhuo and Hu, Shengding and Thai, Zhen and Shen, Junhao and Hu, Jinyi and Han, Xu and Huang, Yujie and Zhang, Yuxiang and others},
  booktitle={Proceedings of the 62nd Annual Meeting of the Association for Computational Linguistics (Volume 1: Long Papers)},
  pages={3828--3850},
  year={2024}
}

@article{rlhf,
 author = {Ouyang, Long and Wu, Jeffrey and Jiang, Xu and Almeida, Diogo and Wainwright, Carroll and Mishkin, Pamela and Zhang, Chong and Agarwal, Sandhini and Slama, Katarina and Ray, Alex and Schulman, John and Hilton, Jacob and Kelton, Fraser and Miller, Luke and Simens, Maddie and Askell, Amanda and Welinder, Peter and Christiano, Paul F and Leike, Jan and Lowe, Ryan},
 booktitle = {Advances in Neural Information Processing Systems},
 editor = {S. Koyejo and S. Mohamed and A. Agarwal and D. Belgrave and K. Cho and A. Oh},
 pages = {27730--27744},
 publisher = {Curran Associates, Inc.},
 title = {Training language models to follow instructions with human feedback},
 url = {https://proceedings.neurips.cc/paper_files/paper/2022/file/b1efde53be364a73914f58805a001731-Paper-Conference.pdf},
 volume = {35},
 year = {2022}
}

@inproceedings{liu2025visual,
  title={Visual-rft: Visual reinforcement fine-tuning},
  author={Liu, Ziyu and Sun, Zeyi and Zang, Yuhang and Dong, Xiaoyi and Cao, Yuhang and Duan, Haodong and Lin, Dahua and Wang, Jiaqi},
  booktitle={Proceedings of the IEEE/CVF International Conference on Computer Vision},
  pages={2034--2044},
  year={2025}
}

@inproceedings{phybench,
title={{PHYB}ench: Holistic Evaluation of Physical Perception and Reasoning in Large Language Models},
author={Shi Qiu and Shaoyang Guo and Zhuo-Yang Song and Yunbo Sun and Zeyu Cai and Jiashen Wei and Tianyu Luo and Yixuan Yin and Zhang Haoxu and Yi Hu and Chenyang Wang and Chencheng Tang and Haoling Chang and Qi Liu and Ziheng Zhou and Tianyu Zhang and Jingtian Zhang and Zhangyi Liu and Minghao Li and Yuku Zhang and Boxuan Jing and Xianqi Yin and Yutong Ren and Zizhuo Fu and Jiaming Ji and Weike Wang and Xudong Tian and Anqi Lv and Laifu Man and Jianxiang Li and Feiyu Tao and Qihua Sun and Zhou Liang and Yushu Mu and Zhongxuan Li and Jing-Jun Zhang and Shutao Zhang and Xiaotian Li and Xingqi Xia and Jiawei Lin and Zheyu Shen and Jiahang Chen and Qiuhao Xiong and Binran Wang and Fengyuan Wang and Niziyang and Bohan Zhang and Fan Cui and shaochangkun and Qing-Hong Cao and Ming-xing Luo and Muhan Zhang and Hua Xing Zhu},
booktitle={The Thirty-ninth Annual Conference on Neural Information Processing Systems Datasets and Benchmarks Track},
year={2025},
url={https://openreview.net/forum?id=brG8FPq1cf}
}

@article{gptoss,
  title={gpt-oss-120b \& gpt-oss-20b model card},
  author={Agarwal, Sandhini and Ahmad, Lama and Ai, Jason and Altman, Sam and Applebaum, Andy and Arbus, Edwin and Arora, Rahul K and Bai, Yu and Baker, Bowen and Bao, Haiming and others},
  journal={arXiv preprint arXiv:2508.10925},
  year={2025}
}

@inproceedings{vllm,
  title={Efficient Memory Management for Large Language Model Serving with PagedAttention},
  author={Woosuk Kwon and Zhuohan Li and Siyuan Zhuang and Ying Sheng and Lianmin Zheng and Cody Hao Yu and Joseph E. Gonzalez and Hao Zhang and Ion Stoica},
  booktitle={Proceedings of the ACM SIGOPS 29th Symposium on Operating Systems Principles},
  year={2023}
}

@article{horawalavithana2023scitune,
  title={Scitune: Aligning large language models with scientific multimodal instructions},
  author={Horawalavithana, Sameera and Munikoti, Sai and Stewart, Ian and Kvinge, Henry},
  journal={arXiv preprint arXiv:2307.01139},
  year={2023}
}

@article{chu2025sft,
  title={Sft memorizes, rl generalizes: A comparative study of foundation model post-training},
  author={Chu, Tianzhe and Zhai, Yuexiang and Yang, Jihan and Tong, Shengbang and Xie, Saining and Schuurmans, Dale and Le, Quoc V and Levine, Sergey and Ma, Yi},
  journal={arXiv preprint arXiv:2501.17161},
  year={2025}
}

@article{shen2025vlm,
  title={Vlm-r1: A stable and generalizable r1-style large vision-language model},
  author={Shen, Haozhan and Liu, Peng and Li, Jingcheng and Fang, Chunxin and Ma, Yibo and Liao, Jiajia and Shen, Qiaoli and Zhang, Zilun and Zhao, Kangjia and Zhang, Qianqian and others},
  journal={arXiv preprint arXiv:2504.07615},
  year={2025}
}

@article{wang2025vl,
  title={Vl-rethinker: Incentivizing self-reflection of vision-language models with reinforcement learning},
  author={Wang, Haozhe and Qu, Chao and Huang, Zuming and Chu, Wei and Lin, Fangzhen and Chen, Wenhu},
  journal={arXiv preprint arXiv:2504.08837},
  year={2025}
}

@misc{anthropic2026claudehaiku55,
      title={Introducing {Claude} {Haiku} 5.5},
      author={{Anthropic}},
      year={2026},
      month={October},
      howpublished={\url{https://www.anthropic.com/claude-haiku-5-5}},
      note={Accessed: 2026-10-12},
}

@article{zhou2025disco,
  title={Disco balances the scales: Adaptive domain-and difficulty-aware reinforcement learning on imbalanced data},
  author={Zhou, Yuhang and Zhu, Jing and Qian, Shengyi and Zhao, Zhuokai and Wang, Xiyao and Liu, Xiaoyu and Li, Ming and Xu, Paiheng and Ai, Wei and Huang, Furong},
  journal={arXiv preprint arXiv:2505.15074},
  year={2025}
}

@article{dai2026qoq,
  title={Qoq-med: Building multimodal clinical foundation models with domain-aware grpo training},
  author={Dai, David and Chen, Peilin and Ekbote, Chanakya and Liang, Paul},
  journal={Advances in Neural Information Processing Systems},
  volume={38},
  pages={37406--37453},
  year={2026}
}

@inproceedings{zhao2026learning,
  title={Learning to reason without external rewards},
  author={Zhao, Xuandong and Kang, Zhewei and Feng, Aosong and Levine, Sergey and Song, Dawn},
  booktitle={International Conference on Learning Representations},
  volume={2026},
  pages={2548--2581},
  year={2026}
}

@article{guo2026humanity,
  title={Humanity's Sixth Sense: Benchmarking Intuitive Visual Reasoning in Multimodal Models},
  author={Guo, Xingang and Gu, Jing and Jang, Brian and Wang, Renxiong and Tyagi, Utkarsh and Quigley, Daniel and Li, Steven and Yan, David and Zhang, Daniel Yue and Yi, Darvin and others},
  journal={arXiv preprint arXiv:2610.08966},
  year={2026}
}

@inproceedings{gunjal2026rubrics,
  title={Rubrics as rewards: Reinforcement learning beyond verifiable domains},
  author={Gunjal, Anisha and Wang, Anthony and Lau, Elaine and Nath, Vaskar and He, Yunzhong and Liu, Bing and Hendryx, Sean},
  booktitle={International Conference on Learning Representations},
  volume={2026},
  pages={127924--127945},
  year={2026}
}

@article{viswanathan2026checklists,
  title={Checklists are better than reward models for aligning language models},
  author={Viswanathan, Vijay and Sun, Yanchao and Kong, Xiang and Cao, Meng and Neubig, Graham and Wu, Sherry},
  journal={Advances in Neural Information Processing Systems},
  volume={38},
  pages={114728--114754},
  year={2026}
}

@inproceedings{chen2026rm,
  title={Rm-r1: Reward modeling as reasoning},
  author={Chen, Xiusi and Li, Gaotang and Wang, Ziqi and Jin, Bowen and Qian, Cheng and Wang, Yu and Wang, Hongru and Zhang, Yu and Zhang, Denghui and Zhang, Tong and others},
  booktitle={International Conference on Learning Representations},
  volume={2026},
  pages={88313--88342},
  year={2026}
}
